%% file: _main.tex
\documentclass[letterpaper,10pt,conference]{formatting/ieeeconf}
\IEEEoverridecommandlockouts
\usepackage{amsmath,amssymb}
\makeatletter
\def\input@path{{formatting/algpseudocodex/}{formatting/fifo-stack/}{formatting/tabto/}{formatting/totcount/}}
\makeatother
\usepackage[noEnd=true,indLines=true]{algpseudocodex}
\usepackage{booktabs}
\usepackage{multirow}
\usepackage{cite}
\usepackage{graphicx}
\usepackage[caption=false,font=footnotesize]{subfig}
\usepackage{float}
\usepackage{placeins}
\usepackage{tabularx}
\usepackage{tikz}
\usepackage{xcolor}
\usepackage{url}
\usepackage{xspace}
\usetikzlibrary{arrows.meta,calc,decorations.pathreplacing,positioning,shapes.geometric}
\usepackage{comment}
\let\labelindent\relax
\usepackage{enumitem}
\setlist{leftmargin=3.5mm}

\floatstyle{ruled}
\newfloat{algorithm}{tbp}{loa}
\floatname{algorithm}{Algorithm}
\def\rootrecycleplan{\textsc{ST-pRRTC-RootRecycle}\xspace}
\def\intervalrootplan{\textsc{IntervalRoot}\xspace}
\def\gpuextend{\textsc{Extend}\xspace}
\algrenewcommand\alglinenumber[1]{\scriptsize #1:}
\algrenewcommand\algorithmicindent{1.0em}
\newcommand{\method}{ST-pRRTC}

\IfFileExists{generated/simulation-count-macros.tex}{
  \input{generated/simulation-count-macros.tex}
}{}

\title{\LARGE ST-pRRTC: Parallel Space-Time RRT-C
with Adaptive Goal-Time Forests}
\author{Duo Zhang\qquad Jintong Li\qquad Junshan Huang\qquad Jingjin Yu%
\thanks{D. Zhang, J. Li, J. Huang, and J. Yu are with the Department of Computer
Science, Rutgers, the State University of New Jersey, Piscataway, NJ, USA.
E-Mails: \texttt{\{duo.zhang, jintong.li, junshan.huang, jingjin.yu\}@rutgers.edu}.}}

\input{figures/figure_hardware}

\begin{document}
\maketitle
\thispagestyle{empty}
\pagestyle{empty}

\begin{abstract}
We propose \method, a GPU-parallel space-time RRT-Connect motion planner for
problems with known obstacle trajectories and unspecified arrival time.
Searching over many arrival times broadens temporal coverage but divides a finite
planning budget among more backward trees. To address the challenge, ST-pRRTC builds a shared forward tree and an adaptive forest of backward goal-time trees.
Its \emph{interval root} formulation samples goal arrival times continuously
and guarantees probabilistic completeness and asymptotic arrival-time optimality
under the stated assumptions in a bounded time domain.
The practical \emph{root recycling} policy has no such guarantees. It adapts
a fixed number of backward trees, replacing later roots while retaining
useful search progress. Experiments on three dynamic benchmarks show that both
variants achieve lower mean first-solution times and earlier mean final arrivals
than ST-RRT* and SI-RRT on problems solved by all compared methods.
Further experiments demonstrate the benefit of recycling over broad
arrival-time ranges.
Real-robot demonstrations show root-recycling ST-pRRTC planning motions for a
UR5e among moving Crazyflie quadrotors.
\end{abstract}

\section{Introduction}\label{sec:intro}
\input{texs/00-intro}

% Queue the interval-root figure before page 2 is composed.
\begin{figure*}[t]
  \centering
  \input{figures/figure_interval_root}
  \caption{Interval root with $N=3$ backward-tree groups for one goal.
  Outlined boxes divide the full arrival range equally; filled bands divide the
  total guided-window duration equally. Matching colors identify groups;
  guided portions need not align with a group's uniform interval.
  The marked samples $T$ lie between grid checks. After a better incumbent $U$,
  both sampling allocations update, while retained roots keep their original
  times and group colors; gray roots are inactive.}
  \label{fig:method-interval}
\end{figure*}

\section{Related Work}\label{sec:related}
\input{texs/03-prelim}

% Queue the root-recycling figure before page 3 is composed.
\begin{figure*}[t]
  \centering
  \input{figures/figure_method}
  \vspace{3mm}
  \caption{Root recycling redirects finite tree capacity toward earlier arrivals.
  A solution arriving at $U$ makes later roots unable to improve arrival time.
  Their slots are reused at earlier collision-free times; hollow circles are
  replacement roots with no inherited nodes. The shared forward tree and earlier
  backward trees are retained. All panels use the same schematic space--time
  coordinates; the root at $U$ remains stored but inactive.}
  \label{fig:method}
\end{figure*}

\section{Problem Formulation}\label{sec:problem}
\input{texs/04-problem}

\section{Space-Time pRRTC}\label{sec:algorithm}
\input{texs/09-algorithm}

\section{Evaluation}\label{sec:evaluation}
\input{texs/12-evaluation}

\section{Conclusion}\label{sec:conclusion}
\input{texs/15-conclusion}
% Keep the remaining tables ahead of the references.
\FloatBarrier

{\small
\bibliographystyle{formatting/IEEEtran}
\bibliography{bib/st_prrtc}
}
\end{document}

%% file: figures/figure_hardware.tex
\newcommand{\hardwaremontage}[5][1]{%
  \begingroup
  \pgfmathsetmacro{\hardwareOverlayWidth}{2.99/(3+(#1))}
  \pgfmathsetmacro{\hardwareSideWidth}{0.99-\hardwareOverlayWidth}
  \pgfmathsetmacro{\hardwareHeight}{3*\hardwareSideWidth+0.02}
  \begin{tikzpicture}[
    x=\linewidth,y=\linewidth,
    inner sep=0pt,outer sep=0pt,
    photo/.style={anchor=south west},
    view number/.style={
      anchor=north west,fill=black!70,text=white,
      font=\scriptsize\bfseries,inner sep=1.3pt}]
    \path[use as bounding box] (0,0) rectangle (1,{\hardwareHeight+0.0525});
    \node[font=\footnotesize,anchor=south]
      at ({\hardwareOverlayWidth/2},{\hardwareHeight+0.0175})
      {Temporal overlay};
    \node[font=\footnotesize,anchor=south]
      at ({\hardwareOverlayWidth+0.01+\hardwareSideWidth/2},{\hardwareHeight+0.0175})
      {Side views};
    \node[photo] at (0,0)
      {\includegraphics[width=\hardwareOverlayWidth\linewidth]{figures/#2}};
    \node[photo] at ({\hardwareOverlayWidth+0.01},{2*(\hardwareSideWidth+0.01)})
      {\includegraphics[width=\hardwareSideWidth\linewidth]{figures/#3}};
    \node[photo] at ({\hardwareOverlayWidth+0.01},{\hardwareSideWidth+0.01})
      {\includegraphics[width=\hardwareSideWidth\linewidth]{figures/#4}};
    \node[photo] at ({\hardwareOverlayWidth+0.01},0)
      {\includegraphics[width=\hardwareSideWidth\linewidth]{figures/#5}};
    \node[view number]
      at ({\hardwareOverlayWidth+0.02},{\hardwareHeight-0.01}) {1};
    \node[view number]
      at ({\hardwareOverlayWidth+0.02},{2*\hardwareSideWidth}) {2};
    \node[view number]
      at ({\hardwareOverlayWidth+0.02},{\hardwareSideWidth-0.01}) {3};
  \end{tikzpicture}%
  \endgroup
}

%% file: texs/00-intro.tex
% Keep the abstract first in the left column; float the montage to the right.
\suppressfloats[t]
\begin{figure}[t]
  \centering
  \hardwaremontage[2049/2475]{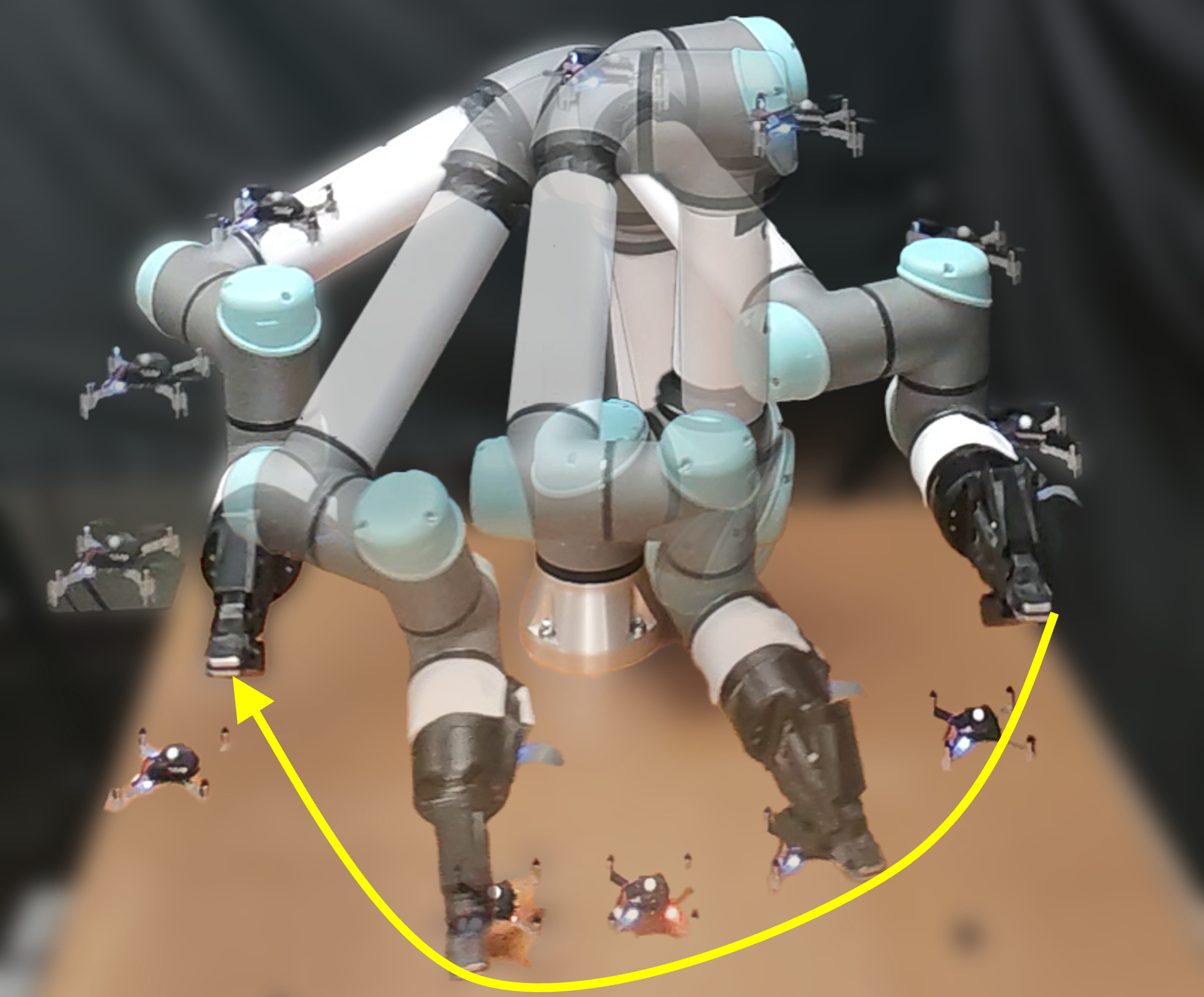}
    {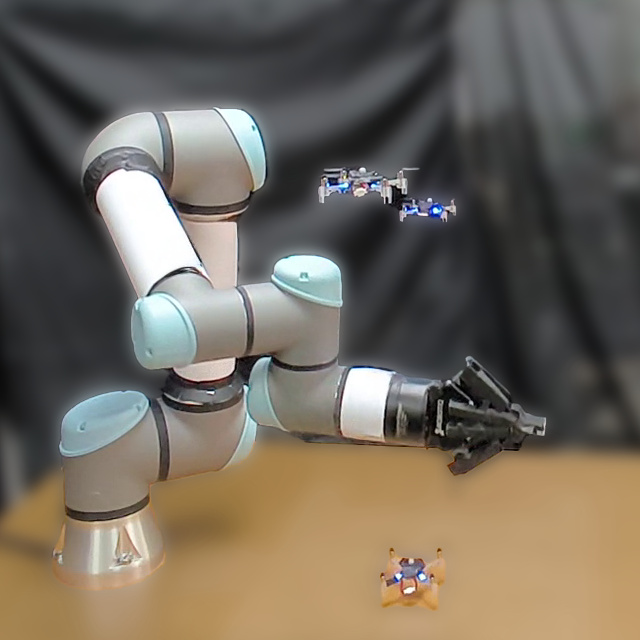}
    {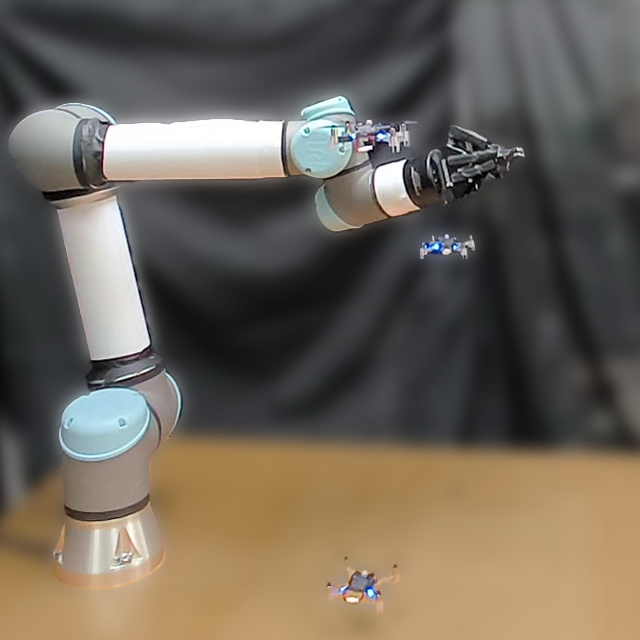}
    {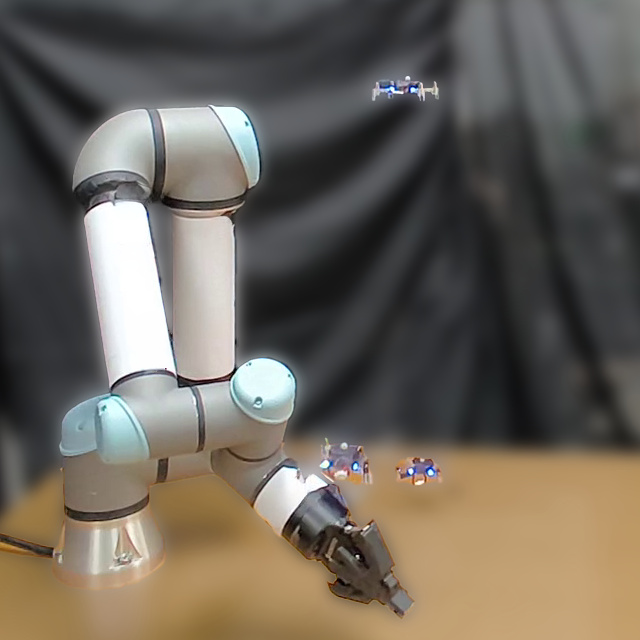}
  \caption{A physical UR5e executes a motion planned by \method{} among three
  moving Crazyflies in the slow demonstration. The large image overlays robot
  and drone poses over time; the yellow arrow indicates end-effector motion toward
  the goal. The right column shows selected side-camera snapshots.}
  \label{fig:hardware-intro}
\end{figure}

Motion planning in shared workspaces requires coordinating robots' movements to reach their goals collision-free. 
Automated warehouses~\cite{lehouxlebacque2024warehouses}, multi-robot assembly cells~\cite{hartmann2023assembly}, and dual-arm tabletop rearrangement~\cite{zhang2025sdar} depend on this coordination to maintain safety and throughput. 
A configuration may be free at one time and blocked at another, requiring joint reasoning over routes and timing, even when obstacle trajectories are known. 
Delays can postpone subsequent tasks, making short execution times essential.

Bidirectional planners such as RRT-Connect~\cite{kuffner2000rrtconnect} explore
from both the start and the goal; pRRTC~\cite{huang2026prrtc} accelerates this
search through GPU batching. With moving obstacles, however, each backward root
requires an arrival time in addition to a goal configuration.
ST-RRT*~\cite{grothe2022strrt} handles unknown arrival times through progressive
time expansion and rewiring. SI-RRT~\cite{kerimov2025sirrt} combines safe intervals
with bidirectional RRT-Connect, initializing backward roots at the upper bounds
of collision-free time intervals at the goal. GPU parallelism enables exploration of many arrival
hypotheses, but divides finite resources among more backward trees, creating a
tradeoff between temporal coverage and individual tree growth. As solutions
improve, search effort should shift toward earlier arrivals while retaining
useful trees.

We propose \method{}, a GPU-parallel space-time RRT-Connect planner built on
pRRTC, to minimize the arrival time given known obstacle trajectories.
Figure~\ref{fig:hardware-intro} illustrates the problem for a 6-DoF arm among
moving aerial obstacles. \method{} brings the following key technical contributions:
\begin{itemize}
  \item \textbf{Parallel Space--Time Search.} \method{} couples a \emph{shared forward
  tree} with a \emph{backward goal-time forest}, exploring routes and timing through
  batched expansion and dynamic collision checking.
  \item \textbf{Continuous Arrival-Time Exploration.} The \emph{interval root}
  formulation samples arrival times continuously within persistent backward tree
  groups. For Algorithm~\ref{alg:interval-root}, we establish probabilistic
  completeness and asymptotic arrival-time optimality under the assumptions
  in Sec.~\ref{sec:analysis} for a bounded arrival-time domain.
  \item \textbf{Adaptive Root Recycling.} A practical policy reuses fixed
  backward-tree slots, replacing roots later than the best known arrival with
  earlier ones while retaining the forward tree and useful backward trees.
  Algorithm~\ref{alg:root-recycle} has no such guarantees; its benefits are empirical.
\end{itemize}

Experiments on Disc2D, Panda-spheres, and tabletop rearrangement show that \method{} variants achieve lower mean first-solution planning times and earlier
mean final arrivals than ST-RRT* and SI-RRT on problems solved by all compared
methods across all tested budgets. Seed repeats support these findings;
ablations demonstrate recycling's benefit over broad arrival-time ranges with
limited tree capacity. Physical UR5e demonstrations show execution among moving
Crazyflies.

%% file: figures/figure_interval_root.tex
\begingroup
\definecolor{intervalgroupone}{RGB}{0,114,178}
\definecolor{intervalgrouptwo}{RGB}{230,159,0}
\definecolor{intervalgroupthree}{RGB}{180,92,160}

\newcommand{\intervalgroupkey}[2]{%
  \raisebox{-0.15ex}{\textcolor{#1}{\rule{1.5ex}{1.5ex}}}\hspace{0.4em}#2}
\newcommand{\intervaluniform}[4]{%
  \draw[draw=#3,fill=#3!5,line width=0.7pt]
    ({#1},24) rectangle ({#2},28);
  \node[text=#3!75!black] at ({(#1+#2)/2},26) {\textbf{#4}};}
\newcommand{\intervalguided}[3]{%
  \fill[#3] ({#1},14) rectangle ({#2},18);}

% Coordinates measure time relative to L_g, in units of the grid spacing.
% Goal checks pass at 1, 4, 5, 6, 9, 10, giving windows
% [0.5,1.5], [3.5,6.5], [8.5,10.5]. There are three groups.
\newcommand{\intervalsamplingpanel}[1]{%
  \begin{tikzpicture}[
    x=0.055\linewidth,y=1mm,font=\footnotesize,
    line cap=round,line join=round]
    \path[use as bounding box] (-5.7,-2) rectangle (12.4,34);
    \ifnum#1=0
      \def\activeupper{12}
      \def\sampletime{4.25}
    \else
      \def\activeupper{9.5}
      \def\sampletime{6.05}
      \fill[black!5] (9.5,4.8) rectangle (12,29);
      \node[text=black!45,font=\scriptsize] at (10.75,26) {inactive};
    \fi

    \foreach \boundary in {0,12}
      \draw[black!15,line width=0.4pt] (\boundary,3.5) -- (\boundary,29);
    \foreach \divider in {1,2}
      \draw[black!18,densely dotted,line width=0.45pt]
        ({\divider*\activeupper/3},12) -- ({\divider*\activeupper/3},28);

    \node[anchor=west] at (-5.7,26) {Uniform intervals};
    \node[anchor=west] at (-5.7,21) {Goal checks};
    \node[anchor=west] at (-5.7,16) {Guided windows};
    \node[anchor=west] at (-5.7,8.5) {Existing roots};

    \ifnum#1=0
      % BEGIN BEFORE SAMPLING: uniform width 4; guided duration 2 per group.
      \intervaluniform{0}{4}{intervalgroupone}{1}
      \intervaluniform{4}{8}{intervalgrouptwo}{2}
      \intervaluniform{8}{12}{intervalgroupthree}{3}
      \intervalguided{0.5}{1.5}{intervalgroupone}
      \intervalguided{3.5}{4.5}{intervalgroupone}
      \intervalguided{4.5}{6.5}{intervalgrouptwo}
      \intervalguided{8.5}{10.5}{intervalgroupthree}
      % END BEFORE SAMPLING
      \draw[black!50,line width=0.45pt] (8.5,14) rectangle (10.5,18);
    \else
      % BEGIN AFTER SAMPLING: U=9.5; guided duration 5/3 per group.
      \intervaluniform{0}{19/6}{intervalgroupone}{1}
      \intervaluniform{19/6}{19/3}{intervalgrouptwo}{2}
      \intervaluniform{19/3}{9.5}{intervalgroupthree}{3}
      \intervalguided{0.5}{1.5}{intervalgroupone}
      \intervalguided{3.5}{25/6}{intervalgroupone}
      \intervalguided{25/6}{35/6}{intervalgrouptwo}
      \intervalguided{35/6}{6.5}{intervalgroupthree}
      \intervalguided{8.5}{9.5}{intervalgroupthree}
      % END AFTER SAMPLING
      \draw[black!50,line width=0.45pt] (8.5,14) rectangle (9.5,18);
    \fi
    \draw[black!50,line width=0.45pt] (0.5,14) rectangle (1.5,18);
    \draw[black!50,line width=0.45pt] (3.5,14) rectangle (6.5,18);

    \foreach \check in {1,4,5,6,9,10} {
      \ifdim\check pt>\activeupper pt
        \def\checkcolor{black!25}
      \else
        \def\checkcolor{black!80}
      \fi
      \fill[\checkcolor] (\check,21) circle[radius=0.55mm];
    }
    \foreach \check in {0,2,3,7,8,11,12} {
      \ifdim\check pt>\activeupper pt
        \def\checkcolor{black!25}
      \else
        \def\checkcolor{black!55}
      \fi
      \node[text=\checkcolor] at (\check,21) {$\times$};
    }

    % A guided draw is continuous, rather than limited to the grid markers.
    \node[circle,draw=black,fill=white,inner sep=0pt,minimum size=3.8pt]
      at (\sampletime,16) {};
    \draw[black!70,line width=0.45pt]
      (\sampletime,15.3) -- (\sampletime,13.7);
    \node[anchor=north,font=\scriptsize] at (\sampletime,13.5) {$T$};

    % Root timestamps and owner colors remain unchanged after repartitioning.
    \foreach \terminal/\owner in {
      1.25/intervalgroupone,4.05/intervalgroupone,
      6.2/intervalgrouptwo,9.2/intervalgroupthree,10.25/intervalgroupthree} {
      \ifdim\terminal pt>\activeupper pt
        \def\rootcolor{black!35}
      \else
        \def\rootcolor{\owner}
      \fi
      \fill[\rootcolor] (\terminal,8.5) circle[radius=2pt];
    }
    \ifnum#1=1
      \draw[black!65,dashed,line width=0.75pt] (9.5,4.8) -- (9.5,30);
      \node[above] at (9.5,30) {$U$};
    \fi
    \draw[black!50,line width=0.5pt,-{Latex[length=1.5mm]}]
      (0,3.5) -- (12.3,3.5);
    \foreach \tick in {0,12}
      \draw[black!50,line width=0.5pt] (\tick,2.8) -- (\tick,4.2);
    \node[below] at (0,2.3) {$L_g$};
    \node[below] at (12,2.3) {$H$};
    \node[below,font=\scriptsize] at (6,2.3) {Arrival time};
  \end{tikzpicture}}

\subfloat[Before incumbent update.\label{fig:interval-before}]{%
  \begin{minipage}[t]{.485\textwidth}
    \centering\intervalsamplingpanel{0}
  \end{minipage}}%
\hfill
\subfloat[After incumbent update.\label{fig:interval-after}]{%
  \begin{minipage}[t]{.485\textwidth}
    \centering\intervalsamplingpanel{1}
  \end{minipage}}
\par\vspace{2pt}
{\footnotesize
  \intervalgroupkey{intervalgroupone}{Group 1}\quad
  \intervalgroupkey{intervalgrouptwo}{Group 2}\quad
  \intervalgroupkey{intervalgroupthree}{Group 3}\qquad
  Grid checks: $\bullet$ free\quad $\times$ blocked}
\par
\endgroup

%% file: texs/03-prelim.tex
\textbf{Motion Planning}.
PRM~\cite{kavraki1996prm} constructs reusable roadmaps, while
RRT~\cite{lavalle1998rrt} and RRT-Connect~\cite{kuffner2000rrtconnect}
grow trees for individual planning queries. RRT* and
PRM*~\cite{karaman2011sampling} add asymptotic optimality, and
BIT*~\cite{gammell2020bit} combines batches of samples with heuristic search
to improve anytime performance. The rotation-stacked visibility graph
(RVG)~\cite{zhang2025rvg} provides efficient multi-query planning for
polygonal robots in $SE(2)$.

In dynamic environments, RRT$^{X}$~\cite{otte2016rrtx} supports rapid replanning
by repairing its search graph as obstacles change.
Time-Based RRT~\cite{sintov2014tbrrt} adds time to
tree states for rendezvous planning with specified timing. For known obstacle
trajectories, SIPP~\cite{phillips2011sipp} reasons over collision-free time intervals.
SI-RRT~\cite{kerimov2025sirrt} combines safe intervals with bidirectional
RRT-Connect, initializing backward roots at the upper bounds of safe intervals
at the goal and returning its first solution. ST-RRT*~\cite{grothe2022strrt} samples
space--time states and expands its arrival-time range while optimizing
arrival time.

Optimization-based methods~\cite{zucker2013chomp,kalakrishnan2011stomp,schulman2014trajopt,zhang2024sip}
can incorporate moving obstacles through time-dependent collision costs or
constraints, but local optimization does not generally guarantee global optimality.

\textbf{Hardware Acceleration}.
VAMP~\cite{thomason2024vamp} accelerates forward kinematics and collision checking through CPU SIMD. FCIT*~\cite{wilson2025fcit} exploits vectorized edge
evaluation to search fully connected graphs without nearest-neighbor structures and
retains asymptotic optimality. On GPUs,
cuRobo~\cite{sundaralingam2023curobo} parallelizes inverse kinematics, geometric
planning, and trajectory optimization; cuRoboV2~\cite{sundaralingam2026curobov2}
adds dynamics-aware optimization and depth-fused distance fields.
pRRTC~\cite{huang2026prrtc} maps the RRT-Connect pipeline to GPU batches.
AORRTC~\cite{wilson2025aorrtc} combines RRT-Connect with AO-X cost-bounded
search~\cite{hauser2016aox} and includes a SIMD-accelerated implementation;
pAORRTC~\cite{huang2026paorrtc} studies a GPU implementation.
ST-pRRTC extends GPU-parallel RRT-Connect to known obstacle trajectories,
using adaptive goal-time forests to allocate backward search effort across
candidate arrival times.

%% file: figures/figure_method.tex
\begingroup
\definecolor{methodforward}{RGB}{0,114,178}
\definecolor{methodbackward}{RGB}{0,158,115}
\definecolor{methodaction}{RGB}{213,94,0}
\definecolor{methodinactive}{RGB}{145,145,145}

\tikzset{
  method panel/.style={
    x=0.020\linewidth,y=1mm,font=\footnotesize,
    line cap=round,line join=round},
  forward/.style={draw=methodforward,line width=1.1pt},
  backward/.style={draw=methodbackward,line width=1.1pt},
  inactive/.style={draw=methodinactive,line width=0.8pt,densely dashed},
  root/.style={circle,fill=methodbackward,draw=methodbackward,
    inner sep=0pt,minimum size=4.5pt},
  empty root/.style={root,fill=white,line width=1.1pt,minimum size=5.5pt}}

% All panels share coordinates: retained nodes never move in space or time.
\newcommand{\methodaxes}{%
  \path[use as bounding box] (0,-3) rectangle (50,42);
  \draw[black!45,thin,-{Latex[length=1.5mm]}] (5,31) -- (48.5,31);
  \node[anchor=east] at (3.4,31) {$q_g$};
  \node[anchor=west] at (48.5,31) {$t$};
  \node[below] at (47.5,30.5) {$H$};
  \node[anchor=east] at (3.4,5) {$q_s$};}
\newcommand{\methodforwardtree}{%
  \draw[forward] (5,5) -- (11,8) -- (18,11.5) -- (24,16);
  \draw[forward] (11,8) -- (15,5.8) (18,11.5) -- (22,9);
  \node[circle,fill=methodforward,inner sep=0pt,minimum size=4.5pt]
    at (5,5) {};}
\newcommand{\methodbackwardtree}[2][backward]{%
  \draw[#1] (#2,31) -- (#2-3,26) -- (#2-6,21);
  \draw[#1] (#2-3,26) -- (#2-5,27.5);}
\newcommand{\methodbound}{%
  \draw[methodaction,densely dashed,line width=0.9pt] (32,2) -- (32,33);
  \node[methodaction,below right,inner sep=1.5pt] at (32,30.5) {$U$};}

\subfloat[Search candidate arrival times.\label{fig:method-search}]{%
\begin{minipage}[t]{0.32\textwidth}
\centering
\begin{tikzpicture}[method panel]
  \methodaxes
  \methodforwardtree
  \foreach \x in {14,23,32,40,46} {
    \methodbackwardtree{\x}
    \node[root] at (\x,31) {};
  }
  \draw[methodbackward,decorate,
    decoration={brace,amplitude=1.3mm}] (14,33.5) -- (46,33.5);
  \node[methodbackward!65!black] at (30,37) {candidate goal times};
  \node[methodbackward!65!black,align=center] at (36,16)
    {backward\\forest};
  \node[methodforward!80!black] at (19,0) {shared forward tree};
\end{tikzpicture}
\end{minipage}}%
\hfill
\subfloat[Find a solution arriving at $U$.\label{fig:method-trigger}]{%
\begin{minipage}[t]{0.32\textwidth}
\centering
\begin{tikzpicture}[method panel]
  \fill[black!5] (32,2) rectangle (48,34);
  \methodaxes
  \methodforwardtree
  \foreach \x in {14,23,32} {
    \methodbackwardtree{\x}
    \node[root] at (\x,31) {};
  }
  \foreach \x in {40,46} {
    \methodbackwardtree[inactive]{\x}
    \node[root,fill=methodinactive,draw=methodinactive] at (\x,31) {};
  }
  \methodbound
  \draw[methodaction,line width=1.4pt]
    (5,5) -- (11,8) -- (18,11.5) -- (24,16)
    -- (26,21) -- (29,26) -- (32,31);
  \node[root,fill=methodaction,draw=methodaction] at (32,31) {};
  \node[methodaction] at (25,37) {new best arrival};
  \node[methodaction] at (15,17.5) {solution};
  \node[methodinactive!80!black,align=center] at (40,11)
    {later roots\\cannot improve};
\end{tikzpicture}
\end{minipage}}%
\hfill
\subfloat[Recycle slots to earlier times.\label{fig:method-recycling}]{%
\begin{minipage}[t]{0.32\textwidth}
\centering
\begin{tikzpicture}[method panel]
  \fill[black!5] (32,2) rectangle (48,34);
  \methodaxes
  \methodforwardtree
  \foreach \x in {14,23} {
    \methodbackwardtree{\x}
    \node[root] at (\x,31) {};
  }
  % The incumbent slot remains present and inactive until a strict improvement.
  \methodbackwardtree[inactive]{32}
  \node[root,fill=methodinactive,draw=methodaction] at (32,31) {};
  \methodbound
  % Crosses mark vacated donor positions; arrows map their slots to new roots.
  \foreach \x in {40,46} {
    \draw[methodinactive,line width=0.9pt]
      (\x-0.65,30.3) -- (\x+0.65,31.7)
      (\x-0.65,31.7) -- (\x+0.65,30.3);
  }
  \draw[methodaction,line width=1pt,-{Latex[length=1.6mm]},shorten >=3pt]
    (40,31) .. controls (38.5,35.5) and (29,35.5) .. (27.5,31);
  \draw[methodaction,line width=1pt,-{Latex[length=1.6mm]},shorten >=3pt]
    (46,31) .. controls (43.5,39) and (21,39) .. (18.5,31);
  \foreach \x in {18.5,27.5} {
    \node[empty root] at (\x,31) {};
  }
  \node[methodaction] at (25,40) {reuse the same slots};
  \node[methodbackward!65!black] at (14,18) {retained};
  \node[methodforward!80!black] at (19,0) {same forward tree};
\end{tikzpicture}
\end{minipage}}
\endgroup

%% file: texs/04-problem.tex
Following the space--time formulation of ST-RRT*~\cite{grothe2022strrt}, let $\mathcal C\subset\mathbb R^d$ be the robot's compact configuration space, equipped with the Euclidean metric $D(q_a,q_b)=\lVert q_b-q_a\rVert$.  Moving-obstacle trajectories are known over the planning horizon $\mathcal T_H=[0,H]$. Let $\mathcal C_{\mathrm{free}}(t)$ denote the collision-free configuration space at time $t$. The space--time state space and its collision-free subset are
\begin{equation}
 \mathcal X=\mathcal C\times\mathcal T_H,\qquad
 \mathcal X_{\mathrm{free}}=
 \{(q,t)\in\mathcal X:q\in\mathcal C_{\mathrm{free}}(t)\}.
 \label{eq:space-time-state-space}
\end{equation}
Starting from $x_s=(q_s,0)$, the robot must reach every goal in the finite sequence $\mathcal G=\langle q_g^1,\ldots,q_g^m\rangle$ in the given order. Each goal is assumed collision-free at some time in $\mathcal T_H$. Moving obstacles may repeatedly block and clear a goal, making the choice of arrival times a central planning challenge.

The robot is holonomic with first-order, velocity-limited motion. With speed limit $v_{\max}>0$ measured in the Euclidean metric, travel from $q_a$ to $q_b$ requires at least
\begin{equation}
 \tau_{\min}(q_a,q_b)=D(q_a,q_b)/v_{\max}.
 \label{eq:minimum-travel-time}
\end{equation}
Thus, a directed space--time motion from $x_a=(q_a,t_a)$ to $x_b=(q_b,t_b)$ requires $t_b>t_a$ and $D(q_a,q_b)\leq v_{\max}(t_b-t_a)$. This reachability relation is asymmetric because time cannot run backward.

Formally, an absolutely continuous trajectory $\pi:[0,T]\rightarrow\mathcal C$ is feasible for the goal sequence $\mathcal G$ if $\pi(0)=q_s$ and there exist visit times $0\leq T_1\leq\cdots\leq T_m=T\leq H$ such that
\begin{equation}
 \begin{aligned}
  \pi(T_i)&=q_g^i,\quad \forall i\in\{1,\ldots,m\},\\
  \pi(t)&\in\mathcal C_{\mathrm{free}}(t),\quad \forall t\in[0,T],\\
  \lVert\dot\pi(t)\rVert&\leq v_{\max},\quad \text{for almost every }t\in[0,T].
 \end{aligned}
 \label{eq:trajectory-feasibility}
\end{equation}
These conditions allow waiting and require every goal to be reached in order. Let $\Pi_H$ denote the set of feasible trajectories. The objective is to minimize the arrival time,
\begin{equation}
 T_H^*=\inf_{\pi\in\Pi_H} c(\pi),\qquad c(\pi)=T.
 \label{eq:arrival-objective}
\end{equation}

With preceding segments fixed, planning reduces to reaching the next goal from the previous arrival state. We therefore present algorithms for a single goal $q_g$ ($m=1$), applied successively along $\mathcal G$. Each query resets local time to zero, shifts obstacle trajectories by their absolute start time, and uses the remaining horizon. The theoretical guarantees concern each single-goal query.

The model assumes deterministic obstacle predictions and open-loop planning over $[0,H]$. A trajectory is \emph{robust} if its space--time graph has positive clearance from $(\mathbb R^d\times\mathcal T_H)\setminus\mathcal X_{\mathrm{free}}$, including configuration-space boundaries~\cite{bekris2021asymptotically}. This clearance assumption is used in Sec.~\ref{sec:analysis}.

%% file: texs/09-algorithm.tex
\subsection{GPU-Parallel Space--Time Forest}
ST-pRRTC maintains a shared forward tree $\mathcal T_s$ rooted at $(q_s,0)$
and a backward forest rooted at candidate goal states $(q_g,T)$.
GPU workers expand both concurrently using shared dynamic collision kernels
and atomically publish improved solutions. The incumbent cost $(U,C)$
prioritizes arrival time $U$, with path length $C$ breaking ties.

The variants differ in allocation and local expansion rules. Interval root
maintains $N$ persistent \emph{backward-tree groups} for $q_g$, each sampling
arrival times continuously from assigned intervals and potentially containing
multiple concrete trees (Fig.~\ref{fig:method-interval}).
Root recycling maintains at most $N$ concrete-tree slots for $q_g$, reusing
slots with root times later than $U$ (Fig.~\ref{fig:method}). Both adapt their
allocation when $U$ decreases.

\subsection{Interval Root}\label{sec:interval-root}
For a single-goal query, let $H$ be the remaining horizon and $\Delta t$ the
environment time step. Times below are local to the query; collision checks
use the corresponding absolute times. The travel-time lower bound for $q_g$
and the first kinematically reachable grid index are
\begin{equation}
 L_g=\frac{D(q_s,q_g)}{v_{\max}},
 \qquad k_g^{\min}=\left\lceil L_g/\Delta t\right\rceil.
 \label{eq:goal-time-lower-bound}
\end{equation}
The planner checks $q_g$ at grid times $k\Delta t\in[L_g,H]$.
Each collision-free check marks a centered window of width $\Delta t$,
clipped to $[L_g,H]$; adjacent windows merge into \emph{guided windows}.
These windows guide sampling but do not certify collision-free intervals:
each proposed root is collision-checked at its sampled time.

With incumbent arrival time $U$ ($U=\infty$ initially), let
$\overline U=\min(H,U)$. While $L_g<\overline U$, divide
$[L_g,\overline U)$ into $N$ equal-width intervals
$\{\mathcal I_i\}_{i=1}^N$. Separately, allocate equal total durations of the
guided windows below $\overline U$ to the same groups, skipping gaps.
A group may receive portions of several windows, including portions outside its
uniform interval.

Group $i$ samples $T$ uniformly from its guided portions with probability
$p_{\rm guided}\in[0,1)$ and from $\mathcal I_i$ otherwise, falling back to
$\mathcal I_i$ if its guided portions are empty. Both choices are continuous;
the uniform component preserves proposal support over $[L_g,\overline U)$.
A collision-free $(q_g,T)$ serves as a virtual root; its first successful
extension creates a concrete backward tree in group $i$.

Each worker alternates forward and backward expansion and selects a group
uniformly, including groups without concrete trees. After sampling $T$,
it draws $q_r$ uniformly from $\mathcal C$ and samples
\begin{equation}
 t_r\sim\operatorname{Uniform}\!\left[
 \frac{D(q_s,q_r)}{v_{\max}},
 T-\frac{D(q_r,q_g)}{v_{\max}}\right],
 \label{eq:conditional-query-time}
\end{equation}
rejecting an empty interval, as in conditional space--time
sampling~\cite{grothe2022strrt}. The query state is $x_r=(q_r,t_r)$.
A backward worker uses the virtual root as its sole source when the selected
group has no active tree or with probability $p_{\rm new}\in(0,1)$.
Otherwise, nearest-neighbor (NN) search considers all active group vertices.
A reused tree's unchanged root time determines its solution arrival.
\begin{algorithm}[h]
\begin{small}
\vspace{0.025in}
\begin{algorithmic}[1]
\If{$\mathcal V=\emptyset$}
  \State \Return \textsc{Trapped}
\EndIf
\State $v_n\leftarrow\arg\min_{v\in\mathcal V}\rho_\sigma(x(v),x_r)$
\If{$\rho_\sigma(x(v_n),x_r)=\infty$}
  \State \Return \textsc{Trapped}
\EndIf
\State $\alpha\leftarrow\min\{1,\eta/\rho_\sigma(x(v_n),x_r)\}$
\State $x_{\rm new}\leftarrow x(v_n)+\alpha(x_r-x(v_n))$
\If{the extension leaves $[0,H]$ or is in collision}
  \State \Return \textsc{Trapped}
\EndIf
\State Materialize a backward tree in $\mathcal T$ if $v_n$ is virtual
\State Append $x_{\rm new}$ to the tree containing $v_n$, with parent $v_n$
\State Test the best-ranked bridge from $x_{\rm new}$ to $\overline{\mathcal T}$
\If{a bridge exists, is valid, and improves $(U,C)$}
  \State Atomically publish the path and update $(U,C)$
\EndIf
\State \Return \textsc{Advanced}
\end{algorithmic}
\vspace{0.025in}
\caption{\gpuextend($\mathcal T,\overline{\mathcal T},\mathcal V,\sigma,x_r,(U,C)$)}
\label{alg:extend}
\end{small}
\end{algorithm}

For a vertex $x=(q,t)$ and sample $x_r=(q_r,t_r)$, let $\sigma=1$ for forward
expansion or $-1$ for backward expansion. With
$\Delta=\sigma(t_r-t)$, NN search over the selected vertex set
uses a directed distance $\rho_\sigma$. It is infinite unless
$\Delta>0$ and $D(q,q_r)\le v_{\max}\Delta$; otherwise its value is
\begin{equation}
 \rho_\sigma(x,x_r)=
 \sqrt{\frac{D(q,q_r)^2+v_{\max}^2(t_r-t)^2}{2}}.
 \label{eq:space-time-metric}
\end{equation}
Steering interpolates configuration and time by the same fraction, permitting
speeds up to $v_{\max}$. Bridge candidates satisfying the same speed and
time-direction constraints are ranked by arrival time, then space--time distance.
Extensions and bridges are collision-checked along their interpolations.
Algorithm~\ref{alg:extend} uses range $\eta$ and the supplied source set
$\mathcal V$. Here, $\mathcal T$ is the forward tree or selected backward-tree
group; $\overline{\mathcal T}$ is the active backward forest or forward tree,
respectively.

When $U$ decreases, repartition the uniform intervals and clipped guided
windows among the same $N$ groups (Fig.~\ref{fig:method-interval}).
Trees rooted below $U$ remain active with unchanged timestamps and group
ownership; all others become inactive.

\vspace{3mm}
\begin{algorithm}[h]
\begin{small}
\vspace{0.025in}
\begin{algorithmic}[1]
\State \textbf{if} $q_s=q_g$ and $x_s\in\mathcal X_{\mathrm{free}}$ \textbf{then return} $[x_s]$
\State Initialize $\mathcal T_s$, $N$ groups for $q_g$, and $(U,C)\leftarrow(\infty,\infty)$
\While{budget $B$ remains and $L_g<\min(H,U)$}
  \ForAll{GPU workers in parallel}
    \State Alternate direction; sample $i$, $T$, and $x_r$ as above
    \State Skip this attempt if the root or query sample is rejected
    \If{assigned to $\mathcal T_s$}
      \State Apply \gpuextend{} to $\mathcal T_s$ using its vertices, $\sigma=1$
    \Else
      \State $\mathcal V\leftarrow$ active vertices in group $i$;
        $u\leftarrow\operatorname{Uniform}(0,1)$
      \State $\mathit{use\_new\_root}\leftarrow
        [\mathcal V=\emptyset]\lor[u\leq p_{\rm new}]$
      \State Use $\mathcal V=\{(q_g,T)\}$ if $\mathit{use\_new\_root}$
      \State Apply \gpuextend{} to group $i$ using $\mathcal V$, $\sigma=-1$
    \EndIf
  \EndFor
  \If{$U$ decreased}
    \State Update groups; deactivate trees rooted at $T\geq U$
  \EndIf
\EndWhile
\State \Return the incumbent path, or failure if none was found
\end{algorithmic}
\vspace{0.025in}
\caption{\intervalrootplan($q_s,q_g,N,H,B$)}
\label{alg:interval-root}
\end{small}
\end{algorithm}

\subsection{Theoretical Properties}\label{sec:analysis}
\input{texs/06-analysis}

\section{Practical Root Recycling}\label{sec:root-recycling}
Interval root can materialize many backward trees across its continuous
arrival-time range, leaving fewer expansions per tree under a finite budget.
For each single-goal query, Root recycling bounds this capacity by $N$
concrete-tree slots. It concentrates expansions on retained trees and reuses
slots rooted later than the incumbent.

Let $L_g^{\rm grid}=k_g^{\min}\Delta t$ be the discrete root-time lower bound.
The planner initializes up to $N$ collision-free roots $(q_g,T_i)$ across
$[L_g^{\rm grid},H)$, with $H$ the query's remaining horizon.
When the incumbent arrival time $U$ decreases, roots with $T_i\geq U$ become
inactive, and slots with $T_i>U$ become donors (Fig.~\ref{fig:method}).
A donor receives a collision-free root in $[L_g^{\rm grid},U)$; its old tree
is discarded. The forward tree and roots below $U$ retain their nodes and
timestamps. A root at $U$ becomes a donor only after a later strict improvement.
If no valid replacement is found, the donor remains inactive.

Root recycling uses configuration-distance NN and maximum-speed tree edges:
\begin{equation}
 t_{\rm new}=t_n+\sigma D(q_n,q_{\rm new})/v_{\max}.
 \label{eq:max-speed-extension}
\end{equation}
The configuration step is capped at $\eta$; bridges may use available time
slack while respecting the speed limit and time direction. Edges and bridges
are collision-checked at absolute times. Valid improvements to $(U,C)$ are
published atomically (Algorithm~\ref{alg:root-recycle}).
\vspace{3mm}

\begin{algorithm}[h]
\begin{small}
\vspace{0.025in}
\begin{algorithmic}[1]
\State \textbf{if} $q_s=q_g$ and $x_s\in\mathcal X_{\mathrm{free}}$ \textbf{then return} $[x_s]$
\State Choose up to $N$ collision-free roots for $q_g$ in $[L_g^{\rm grid},H)$
\State Initialize $\mathcal T_s$, the backward forest, and $(U,C)\leftarrow(\infty,\infty)$
\While{budget $B$ remains}
  \ForAll{GPU workers in parallel}
    \State Assign the worker to the start tree or backward forest
    \State Assign backward workers to active roots round-robin; skip if none
    \State Sample $q_r$; propose $x_{\rm new}$ using Eq.~\eqref{eq:max-speed-extension}
    \State Skip if no proposal, $t_{\rm new}\notin[0,H]$, or the edge collides
    \State Append $x_{\rm new}$ to the selected tree
    \State Test a bridge; atomically publish valid improvements to $(U,C)$
  \EndFor
  \If{$U$ decreased}
    \State Deactivate roots with $T_i\geq U$
    \ForAll{root slots $i$ with $T_i>U$}
      \State Search for a collision-free replacement in $[L_g^{\rm grid},U)$
      \If{a collision-free candidate $T_i^{\prime}$ is found}
        \State Invalidate the old generation
        \State Install the empty root $(q_g,T_i^{\prime})$
      \EndIf
    \EndFor
  \EndIf
\EndWhile
\State \Return the incumbent path, or failure if none was found
\end{algorithmic}
\vspace{0.025in}
\caption{\rootrecycleplan($q_s,q_g,N,H,B$)}
\label{alg:root-recycle}
\end{small}
\end{algorithm}

Root recycling has no general completeness or arrival-time optimality guarantee:
initial roots can miss reachable arrival times, and recycling requires an
incumbent improvement. Its finite-budget performance is supported empirically
in Sec.~\ref{sec:evaluation}. Maximum-speed edges can approximate waiting and
slower motion under the first-order model. Alternating velocities
$\pm v_{\max}e$, for a unit direction $e$, with forward fraction
$(1+s/v_{\max})/2$ realizes average velocity $se$ for
$0\le s\le v_{\max}$. Within the clearance of a robust reference trajectory,
sufficiently short cycles approximate this average motion; $s=0$ approximates
waiting.

%% file: texs/06-analysis.tex
For one goal and fixed $H$, assume unlimited sampling and storage, fair worker
scheduling, and sound collision validation that accepts sufficiently short
collision-free edges. Algorithm~\ref{alg:interval-root} is probabilistically
complete for robust trajectories arriving before $H$. If such trajectories
approach $T_H^*$ in Eq.~\eqref{eq:arrival-objective} ($m=1$), its incumbent
arrival converges to $T_H^*$ almost surely.

\emph{Proof sketch.} Fix $b\leq H$ admitting a robust arrival before $b$.
Slight time dilation and a terminal wait create strict speed slack within
the free tube. Cover this reference by finitely many small, temporally ordered
neighborhoods. The uniform root component and conditional state sampling give
each neighborhood a positive sampling probability bounded below while $U\geq b$.
A retained forward vertex in the preceding neighborhood is a reachable NN
candidate; the selected NN is no farther away. For sufficiently small
neighborhoods, its entire extension stays in the free tube and reaches the
sample, adapting the covering argument of~\cite{kleinbort2019completeness}.
Since forward vertices persist, the terminal neighborhood is reached almost
surely unless $U<b$ is found earlier.

A fresh root at $T<b$ then has positive probability of extending backward into
the terminal free tube. All bridges from that extension to the forward tree
have arrival time $T$, so ranking selects the nearest admissible vertex.
The retained terminal vertex bounds this distance, ensuring a valid local bridge.
These opportunities remain bounded away from zero under incumbent updates;
hence $U<b$ is obtained almost surely. Taking $b=H$ gives completeness, and a
countable sequence $b\downarrow T_H^*$ gives almost-sure arrival-time optimality.
Path length remains a tie-breaker.

\begin{comment}
Consider robust near-optimal trajectories arriving before $H$, a sound local
collision checker, and unlimited sampling and storage. For bounded $H$, the
probabilistic-completeness argument of~\cite{kleinbort2019completeness} applies
to Interval root's directed space--time search under our robust-trajectory
assumption. Continuous goal-root sampling provides the exact-goal connection,
while incumbent updates preserve the required vertices and sampling support
below the improving arrival bound.

The first-order dynamics and arrival-time cost are Lipschitz.
Almost-sure asymptotic optimality then follows from the AO-X
cost-bounding argument~\cite{hauser2016aox,kleinbort2020refined}, with bounded-cost completeness
preserved under incumbent updates.

For unbounded arrival times, progressively double the horizon through
$H_0,2H_0,4H_0,\ldots$, while allocating unbounded sampling effort to each
finite horizon. After finding incumbent $U$, search below $U$ suffices.
\end{comment}

%% file: texs/12-evaluation.tex
% Queue the benchmark scenes and success plot before the protocol text.
\begin{figure}[t]
  \centering
  \begingroup
  \setlength{\fboxsep}{0pt}
  \setlength{\fboxrule}{0.4pt}
  \subfloat[Disc2D.\label{fig:benchmark-disc2d}]{%
    \raisebox{\fboxrule}{\fbox{%
      \includegraphics[width=\dimexpr.32\linewidth-2\fboxrule\relax]
        {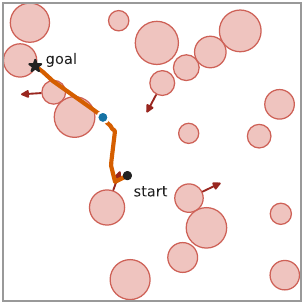}}}}%
  \hfill
  \subfloat[Panda-spheres.\label{fig:benchmark-panda}]{%
    \raisebox{\fboxrule}{\fbox{%
      \includegraphics[width=\dimexpr.32\linewidth-2\fboxrule\relax]
        {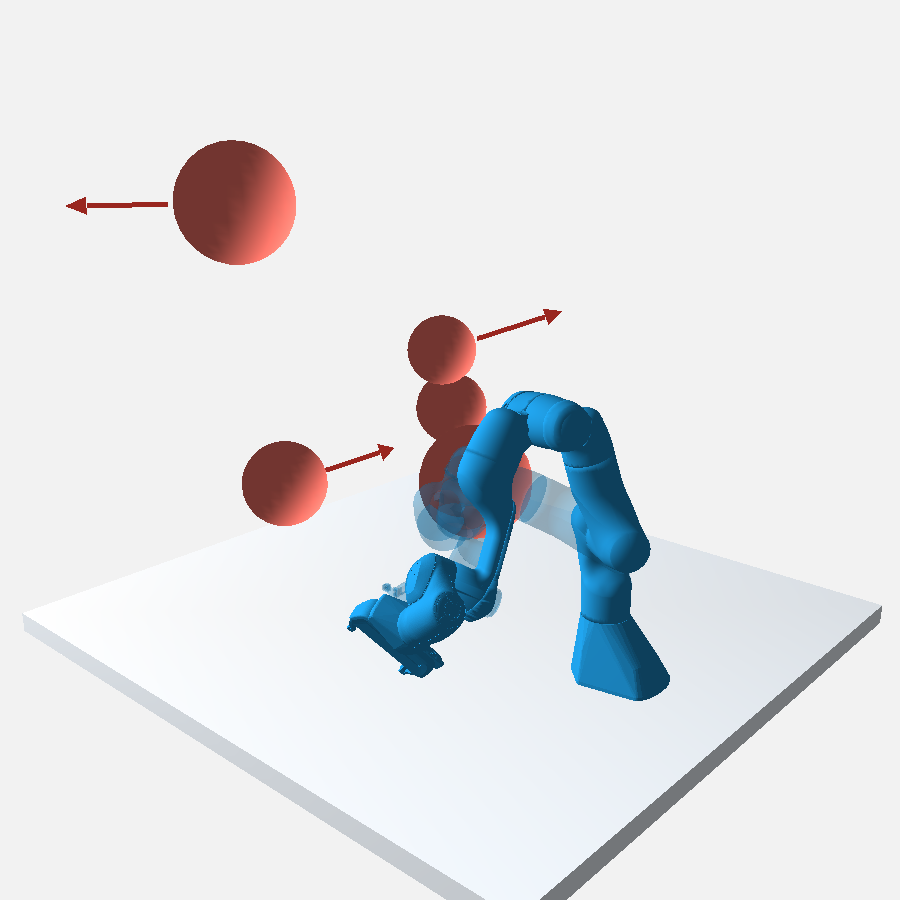}}}}%
  \hfill
  \subfloat[Tabletop.\label{fig:benchmark-tabletop}]{%
    \raisebox{\fboxrule}{\fbox{%
      \includegraphics[width=\dimexpr.32\linewidth-2\fboxrule\relax]
        {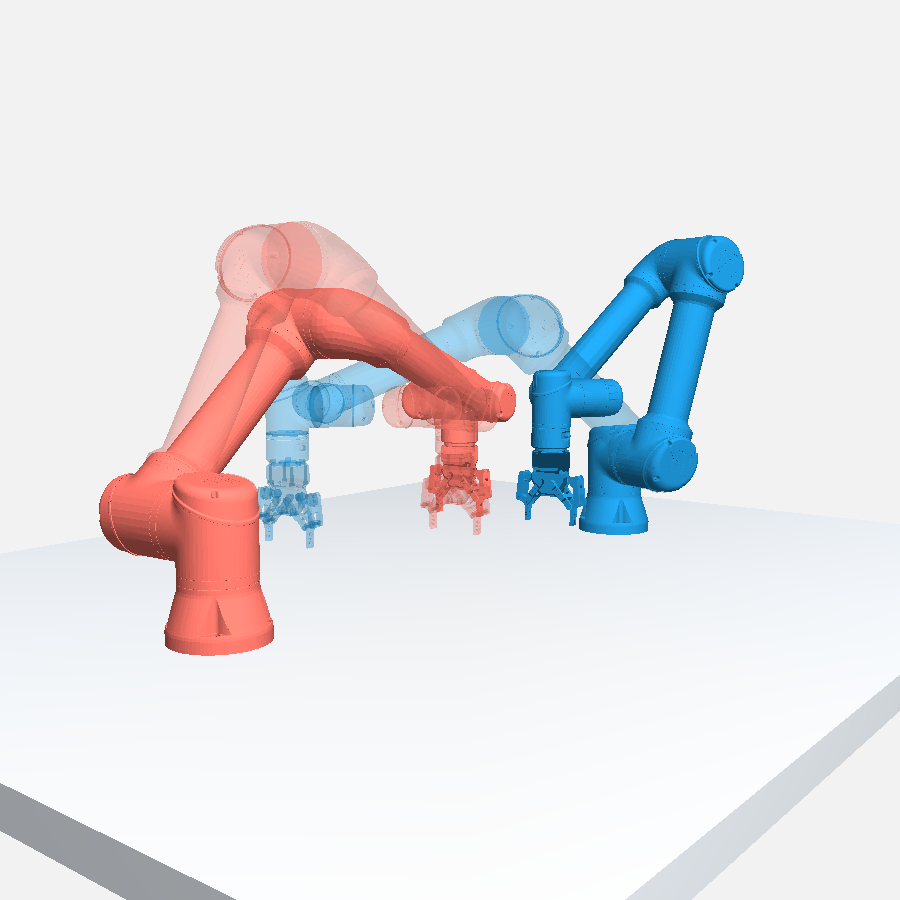}}}}
  \endgroup
  \vspace{3mm}
  \caption{Representative benchmark scenes from root-recycling solutions.
  Blue denotes the planned robot and red the moving obstacles; translucent
  arms indicate goal configurations. Both tabletop arms are shown at their
  start (solid) and goal (translucent) configurations. Arrows in the other
  panels show selected obstacle motion directions. The orange Disc2D curve
  is the time-parameterized robot path. The table is visual context only.}
  \label{fig:benchmark-scenes}
\end{figure}

\begin{figure*}[t]
  \centering
  \raisebox{-6pt}[\height][\depth]{%
    \includegraphics[width=.74\textwidth]{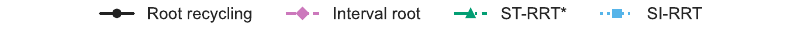}}
  \par\vspace{1pt}
  \subfloat[Disc2D.]{%
    \includegraphics[width=.32\textwidth]{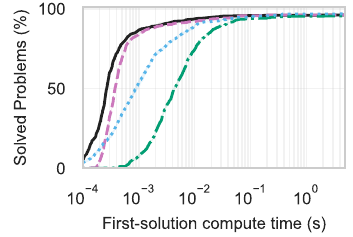}}%
  \hfill
  \subfloat[Panda-spheres.]{%
    \includegraphics[width=.32\textwidth]{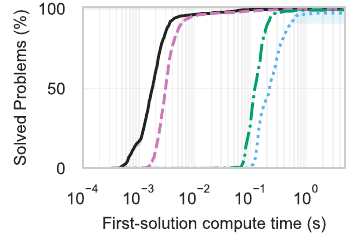}}%
  \hfill
  \subfloat[Tabletop.]{%
    \includegraphics[width=.32\textwidth]{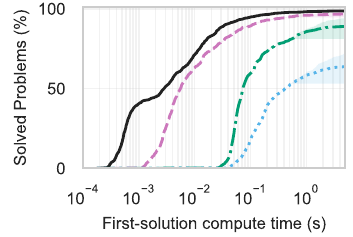}}%
    \vspace{3mm}
  \caption{Cumulative first-solution success. Lines average the four planning
  budgets; shading spans their minimum and maximum. All scenes are included.}
  \label{fig:main-success}
\end{figure*}

\subsection{Simulation Protocol}
We compare Root recycling, Interval root, ST-RRT*, and SI-RRT in the main
study; No recycling is the allocation ablation. SI-RRT
preserves the authors' safe-interval construction, bidirectional scheduling, and
first-solution behavior, while ST-RRT* uses its OMPL implementation
~\cite{sucan2012ompl}.

ST-pRRTC reuses pRRTC's GPU collision checker~\cite{huang2026prrtc}.
Dynamic-obstacle positions are interpolated at each absolute collision-query
time. For Panda-spheres and tabletop, both CPU baselines use Pinocchio with
the HPP-FCL collision backend; Disc2D uses direct disc-overlap tests.

The frozen study contains 300 procedurally generated Disc2D scenes, 300 Panda
scenes with moving spheres, and all 962 tabletop transitions. The tabletop suite
derives from dual-arm rearrangement tasks generated for SDAR~\cite{zhang2025sdar}.
One UR5e is replanned against the prescribed motion of the other arm. The table
and cubes constrain task and IK generation but are omitted from the final
robot-only motion-planning queries. If that transition lasts $T$, the planning
horizon is $H_{\rm scene}=2T$. Figure~\ref{fig:benchmark-scenes} illustrates the
three problem families, and Table~\ref{tab:benchmarks} summarizes their scales.
All suites use a 0.02~s environment time step; $v_{\max}$ is the velocity bound
in the suite metric.

\begin{table}[htbp]
  \centering
  \caption{Benchmark suites.}
  \label{tab:benchmarks}
  \begingroup
  \small
  \setlength{\tabcolsep}{2pt}
  \begin{tabular*}{\linewidth}{@{\extracolsep{\fill}}lclccc@{}}
    \toprule
    Suite & DoF & \shortstack{Moving\\geometry} & Horizon & $v_{\max}$ & Cases \\
    \midrule
    Disc2D & 2 & 20 discs & 8.0~s & 1.2~m/s & 300 \\
    Panda-spheres & 7 & 5 spheres & 6.0~s & 3.14~rad/s & 300 \\
    Tabletop & 6 & other arm & $2T$ & $\pi/2$~rad/s & 962 \\
    \bottomrule
  \end{tabular*}
  \endgroup
\end{table}

The full-dataset comparison uses planning budgets $\{0.5,1,3,5\}$~s and
planner seed 1 for every method;
ST-pRRTC requests 64 backward roots or groups and batches 1024 new
configurations. Root recycling and No recycling retain the original local
extension rule; Interval root uses the space--time core of
Sec.~\ref{sec:interval-root}, with $p_{\rm new}=0.05$ and $p_{\rm guided}=0.95$.

Following pRRTC~\cite{huang2026prrtc}, scene construction and upload are
excluded from timing. We report ST-pRRTC GPU-kernel time in the repeated-query
regime, where one loaded scene serves successive planning queries and setup is
amortized; cuRoboV2 similarly separates GPU-kernel execution from framework and
launch overhead~\cite{sundaralingam2026curobov2}. CPU baselines are timed within
their planning calls after scene construction. Trials run sequentially on an
NVIDIA RTX~4090 (24~GB) and an Intel Core~i9-14900KF; the CPU baselines are
single-threaded.

\subsection{Main Comparison}
\begin{table}[!t]
  \centering
  \caption{Success and first-solution time.}
  \label{tab:compute-success}
  \begingroup
  \footnotesize
  \renewcommand{\arraystretch}{0.98}
  \setlength{\tabcolsep}{2pt}
  \begin{tabular*}{\linewidth}{@{\extracolsep{\fill}}cclrrrr@{}}
  \toprule
   & \shortstack{Budget\\(s)} & Metric & \multicolumn{1}{c}{\shortstack{Root\\recycling}} & \multicolumn{1}{c}{\shortstack{Interval\\root}} & \multicolumn{1}{c}{\shortstack{ST-RRT*}} & \multicolumn{1}{c}{\shortstack{SI-RRT}} \\
  \midrule
  \multirow{8}{*}[-2.25pt]{\rotatebox[origin=c]{90}{\textbf{Disc2D}}} & 0.5 & $S$ (\%) & 95.7 & 96.0 & 95.0 & \textbf{96.7} \\
   & & $\bar t$ (ms) & 1.59 & \textbf{1.31} & 9.85 & 4.66 \\
  \addlinespace[1.5pt]
   & 1 & $S$ (\%) & 96.0 & 96.0 & 95.3 & \textbf{96.7} \\
   & & $\bar t$ (ms) & \textbf{1.17} & 1.37 & 11.40 & 5.62 \\
  \addlinespace[1.5pt]
   & 3 & $S$ (\%) & 96.3 & 96.0 & 95.7 & \textbf{96.7} \\
   & & $\bar t$ (ms) & \textbf{0.93} & 1.34 & 11.48 & 5.58 \\
  \addlinespace[1.5pt]
   & 5 & $S$ (\%) & 96.0 & 96.0 & 96.0 & \textbf{97.0} \\
   & & $\bar t$ (ms) & \textbf{1.28} & 1.43 & 28.71 & 6.91 \\
  \midrule
  \multirow{8}{*}[-2.25pt]{\rotatebox[origin=c]{90}{\textbf{Panda-spheres}}} & 0.5 & $S$ (\%) & \textbf{99.7} & \textbf{99.7} & 98.3 & 90.7 \\
   & & $\bar t$ (ms) & \textbf{3.16} & 5.33 & 130.66 & 227.27 \\
  \addlinespace[1.5pt]
   & 1 & $S$ (\%) & \textbf{99.7} & \textbf{99.7} & 99.3 & 98.7 \\
   & & $\bar t$ (ms) & \textbf{3.36} & 5.26 & 134.28 & 256.67 \\
  \addlinespace[1.5pt]
   & 3 & $S$ (\%) & \textbf{99.7} & \textbf{99.7} & 99.3 & \textbf{99.7} \\
   & & $\bar t$ (ms) & \textbf{3.37} & 5.12 & 135.92 & 267.38 \\
  \addlinespace[1.5pt]
   & 5 & $S$ (\%) & \textbf{99.7} & \textbf{99.7} & \textbf{99.7} & \textbf{99.7} \\
   & & $\bar t$ (ms) & \textbf{3.67} & 6.01 & 135.65 & 266.93 \\
  \midrule
  \multirow{8}{*}[-2.25pt]{\rotatebox[origin=c]{90}{\textbf{Tabletop}}} & 0.5 & $S$ (\%) & \textbf{97.6} & 94.6 & 81.1 & 53.0 \\
   & & $\bar t$ (ms) & \textbf{3.15} & 7.00 & 71.79 & 169.09 \\
  \addlinespace[1.5pt]
   & 1 & $S$ (\%) & \textbf{98.1} & 96.5 & 87.4 & 60.5 \\
   & & $\bar t$ (ms) & \textbf{3.72} & 8.21 & 96.38 & 242.87 \\
  \addlinespace[1.5pt]
   & 3 & $S$ (\%) & \textbf{99.3} & 97.5 & 92.8 & 69.5 \\
   & & $\bar t$ (ms) & \textbf{4.77} & 13.94 & 136.06 & 450.04 \\
  \addlinespace[1.5pt]
   & 5 & $S$ (\%) & \textbf{99.4} & 98.1 & 94.2 & 72.1 \\
   & & $\bar t$ (ms) & \textbf{5.16} & 17.65 & 162.92 & 577.15 \\
  \bottomrule
  \end{tabular*}
  \endgroup
\end{table}

\begin{figure*}[t]
  \centering
  \raisebox{-6pt}[\height][\depth]{%
    \includegraphics[width=.74\textwidth]{figures/simulation-method-legend.pdf}}
  \par\vspace{1pt}
  \subfloat[Disc2D.]{%
    \includegraphics[width=.32\textwidth]{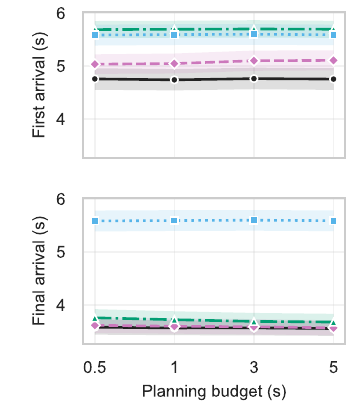}}%
  \hfill
  \subfloat[Panda-spheres.]{%
    \includegraphics[width=.32\textwidth]{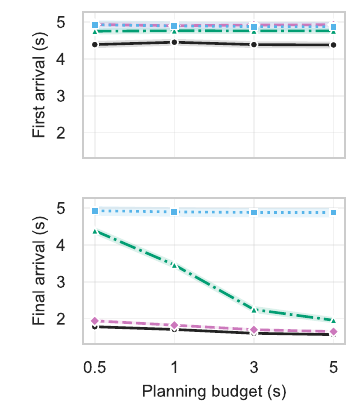}}%
  \hfill
  \subfloat[Tabletop.]{%
    \includegraphics[width=.32\textwidth]{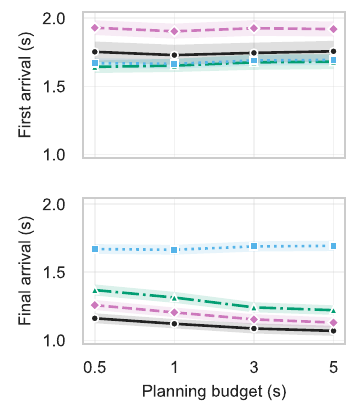}}%
    \vspace{3mm}
  \caption{First-solution (top) and final (bottom) arrival times. Means and
  95\% bootstrap intervals use the common-success set at each budget; the two
  rows share a vertical scale within each benchmark.}
  \label{fig:main-arrival}
\end{figure*}

Planner-reported success and the cumulative first-solution curves use every
scene. First-solution compute time, first and final arrival, and path length use
only cases solved by all four displayed methods at the same benchmark and
budget. At 5~s, the common sets contain 283
Disc2D, 298 Panda-spheres, and 693 tabletop cases. The common-success
set is recomputed for each planning budget.
Path length is Euclidean for Disc2D and joint-space length for both manipulators.

\begin{figure*}[t]
  \centering
  \raisebox{-6pt}[\height][\depth]{%
    \includegraphics[width=.74\textwidth]{figures/simulation-method-legend.pdf}}
  \par\vspace{1pt}
  \subfloat[Disc2D.]{%
    \includegraphics[width=.32\textwidth]{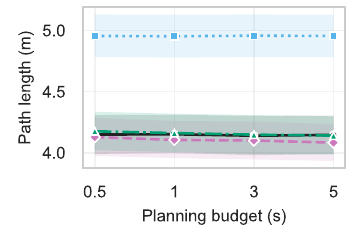}}%
  \hfill
  \subfloat[Panda-spheres.]{%
    \includegraphics[width=.32\textwidth]{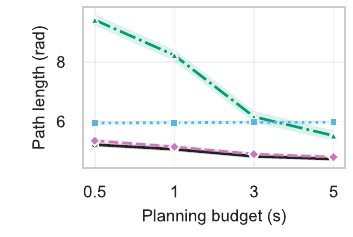}}%
  \hfill
  \subfloat[Tabletop.]{%
    \includegraphics[width=.32\textwidth]{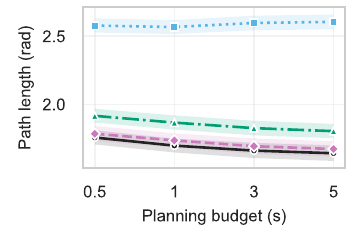}}%
      \vspace{3mm}
  \caption{Final path length on the common-success sets. Shading shows 95\%
  bootstrap intervals; Disc2D uses metres and the manipulators use joint-space
  radians.}
  \label{fig:main-path-length}
\end{figure*}

Figure~\ref{fig:main-success} groups the four
budget-specific cumulative first-solution curves: each line is their mean and
the shaded region spans their minimum and maximum. Figure~\ref{fig:main-arrival} compares first and final arrival
on the same vertical scale within each benchmark, while
Fig.~\ref{fig:main-path-length} reports path length. Table~\ref{tab:compute-success}
lists success rates $S$ over all scenes and mean first-solution compute times
$\bar t$ over the four-method common-success sets, for every budget. Times in
the table are in milliseconds; ST-pRRTC uses GPU-kernel time and the CPU
baselines use planning-call time. Bold entries mark the best value for each
benchmark and budget, including ties.

Across all three benchmarks and tested planning budgets, both ST-pRRTC variants
achieve lower mean first-solution compute times and earlier mean final arrivals
than ST-RRT* and SI-RRT on the common-success sets. Root recycling also achieves
the highest tabletop success at every budget, while remaining competitive on
Disc2D and Panda-spheres.

These results support Root recycling as an effective finite-budget planner.
Its comparison with Interval root reflects both allocation and expansion rules.

% Queue the ablation figure before the seed discussion fills the page.
\begin{figure*}[!t]
  \centering
  \raisebox{-6pt}[\height][\depth]{%
    \includegraphics[width=.74\textwidth,clip]{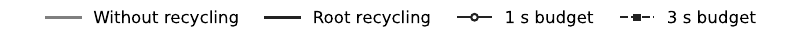}}
  \par\vspace{1pt}
  \subfloat[Initial arrival-time range.\label{fig:capacity-range}]{%
    \includegraphics[width=.32\textwidth,clip]{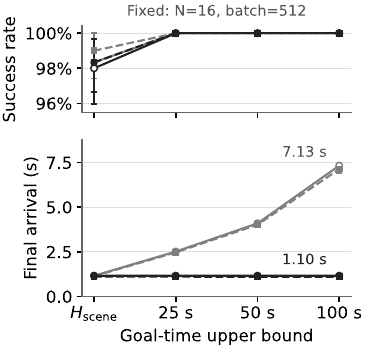}}%
  \hfill
  \subfloat[Number of goal roots.\label{fig:capacity-roots}]{%
    \includegraphics[width=.32\textwidth,clip]{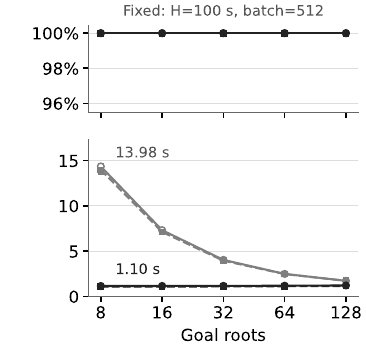}}%
  \hfill
  \subfloat[Configurations per batch.\label{fig:capacity-batch}]{%
    \includegraphics[width=.32\textwidth,clip]{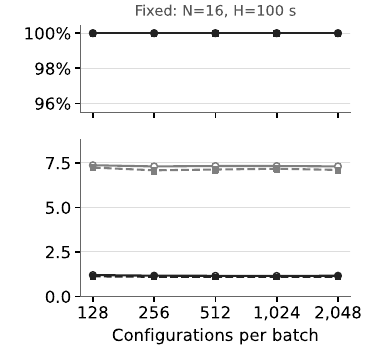}}
      \vspace{3mm}
  \caption{Sensitivity to the initial arrival-time range, root count, and batch
  size on 300 tabletop cases. Each column varies one parameter with the other
  settings fixed as shown. Top: success rate over all cases. Bottom: mean final
  arrival on cases solved by both methods at each setting and budget. Error bars
  are clustered 95\% bootstrap intervals; point labels report 3~s-budget means.
  $H_{\rm scene}$ denotes the per-transition horizon.}
  \label{fig:goal-time-capacity}
    \vspace{3mm}
\end{figure*}

\subsection{Seed Sensitivity}\label{sec:seed-sensitivity}
\begin{table}[!t]
  \centering
  \caption{Seed sensitivity.}
  \label{tab:seed-sensitivity}
  \begingroup
  \footnotesize
  \setlength{\tabcolsep}{1.5pt}
  % Generated by report_icra_seed_sensitivity.py; ranges are not CIs.
  \begin{tabular*}{\linewidth}{@{\extracolsep{\fill}}lrrrrrr@{}}
  \toprule
  & \multicolumn{4}{c}{Solved problems (\%)} & \multicolumn{2}{c}{5-s budget: seed means} \\
  \cmidrule(lr){2-5}\cmidrule(l){6-7}
  Method & 0.5 s & 1 s & 3 s & 5 s & \shortstack{First-solution\\time (ms)} & \shortstack{Final arrival\\(s)} \\
  \midrule
  \multicolumn{7}{l}{\textit{Disc2D} (95 common successes at 5 s)} \\
  Root recycling & \textbf{97--98} & \textbf{97--98} & \textbf{96--98} & \textbf{97--98} & \textbf{1.10--1.92} & \textbf{3.538--3.555} \\
  Interval root & 96--97 & 96--97 & \textbf{97} & 96--97 & 1.26--1.97 & 3.548--3.554 \\
  ST-RRT* & 95 & 95 & 95 & 95--96 & 14.62--17.96 & 3.650--3.724 \\
  SI-RRT & 95--96 & 96--97 & 96--97 & 97 & 3.53--7.43 & 5.414--5.704 \\
  \midrule
  \multicolumn{7}{l}{\textit{Panda-spheres} (99 common successes at 5 s)} \\
  Root recycling & \textbf{100} & \textbf{100} & \textbf{100} & \textbf{100} & \textbf{4.45--5.89} & \textbf{1.547--1.573} \\
  Interval root & \textbf{100} & \textbf{100} & \textbf{100} & \textbf{100} & 8.40--10.51 & 1.630--1.634 \\
  ST-RRT* & 98--99 & 99 & 99 & 99 & 73.19--138.24 & 1.916--2.112 \\
  SI-RRT & 41--93 & 78--100 & \textbf{100} & \textbf{100} & 258.90--703.29 & 3.720--4.714 \\
  \midrule
  \multicolumn{7}{l}{\textit{Tabletop} (63 common successes at 5 s)} \\
  Root recycling & \textbf{95--97} & \textbf{98} & \textbf{99--100} & \textbf{99--100} & \textbf{2.08--3.18} & \textbf{1.061--1.067} \\
  Interval root & 93--94 & 95 & 95--96 & 96--97 & 5.52--6.01 & 1.131--1.134 \\
  ST-RRT* & 77--84 & 86--88 & 91--92 & 92--94 & 102.05--112.62 & 1.204--1.225 \\
  SI-RRT & 51--61 & 58--61 & 64--73 & 69--75 & 344.88--683.81 & 1.796--1.834 \\
  \bottomrule
  \end{tabular*}
  \endgroup
\end{table}

We select 100 problems per suite independently of planner outcomes and use
the same problems for all four methods, budgets, and seeds $\{1,2,3\}$.
Table~\ref{tab:seed-sensitivity} reports the ranges across seeds.
Quality metrics use common successes across all four methods and three seeds. At 5~s, the common sets contain 95 Disc2D, 99 Panda-spheres, and 63 tabletop problems.

Across all three seeds, both ST-pRRTC variants retain lower mean first-solution
compute times and earlier mean final arrivals than both CPU baselines at the
5~s budget. Their success rates remain stable across seeds.

\subsection{Goal-Time Range and Capacity Ablations}

The main-protocol allocation ablation uses the same dense temporal grid with
and without recycling, allowing a fixed forest to reinforce mature trees.
To study allocation across broad goal-time ranges, we use 300 tabletop cases. In
Fig.~\ref{fig:goal-time-capacity}, $H_{\rm scene}$ is the dataset horizon; the
other conditions replace it with common 25, 50, and 100~s upper bounds. The
range sweep uses 16 roots and 512 configurations. The root sweep varies
$\{8,16,32,64,128\}$ at 512 configurations and 100~s, while the batch sweep
varies $\{128,256,512,1024,2048\}$ configurations at 16 roots and 100~s. Both methods solve all 300 cases in
every fixed-horizon condition, including the root-count and batch-size sweeps.
Under $H_{\rm scene}$, success ranges from 98\% to 99\% across methods and
budgets.

At 3~s, widening the fixed 16-root range from 25 to 100~s increases the
no-recycling mean arrival from 2.47 to 7.13~s, while root recycling remains
between 1.10 and 1.11~s. With eight roots over 100~s, both methods solve all 300
cases, but no recycling arrives at 13.98~s on average versus 1.10~s with
recycling. Increasing the root-recycling batch from 128 to 512 configurations
improves mean arrival from 1.13 to 1.10~s; 1024 and 2048 provide no measurable
additional improvement. These ablations highlight the importance of root recycling
when a limited number of roots must cover a broad planning horizon. With
sufficiently dense root coverage, planning without recycling can achieve similar
arrival-time performance.

\subsection{Real-Robot Demonstrations}
\begin{figure*}[t]
  \centering
  \subfloat[Two-Crazyflie top--center--bottom reaching.\label{fig:hardware-two}]{%
    \resizebox{.325\textwidth}{!}{%
    \begin{minipage}[t]{.37\textwidth}
      \centering
      \hardwaremontage{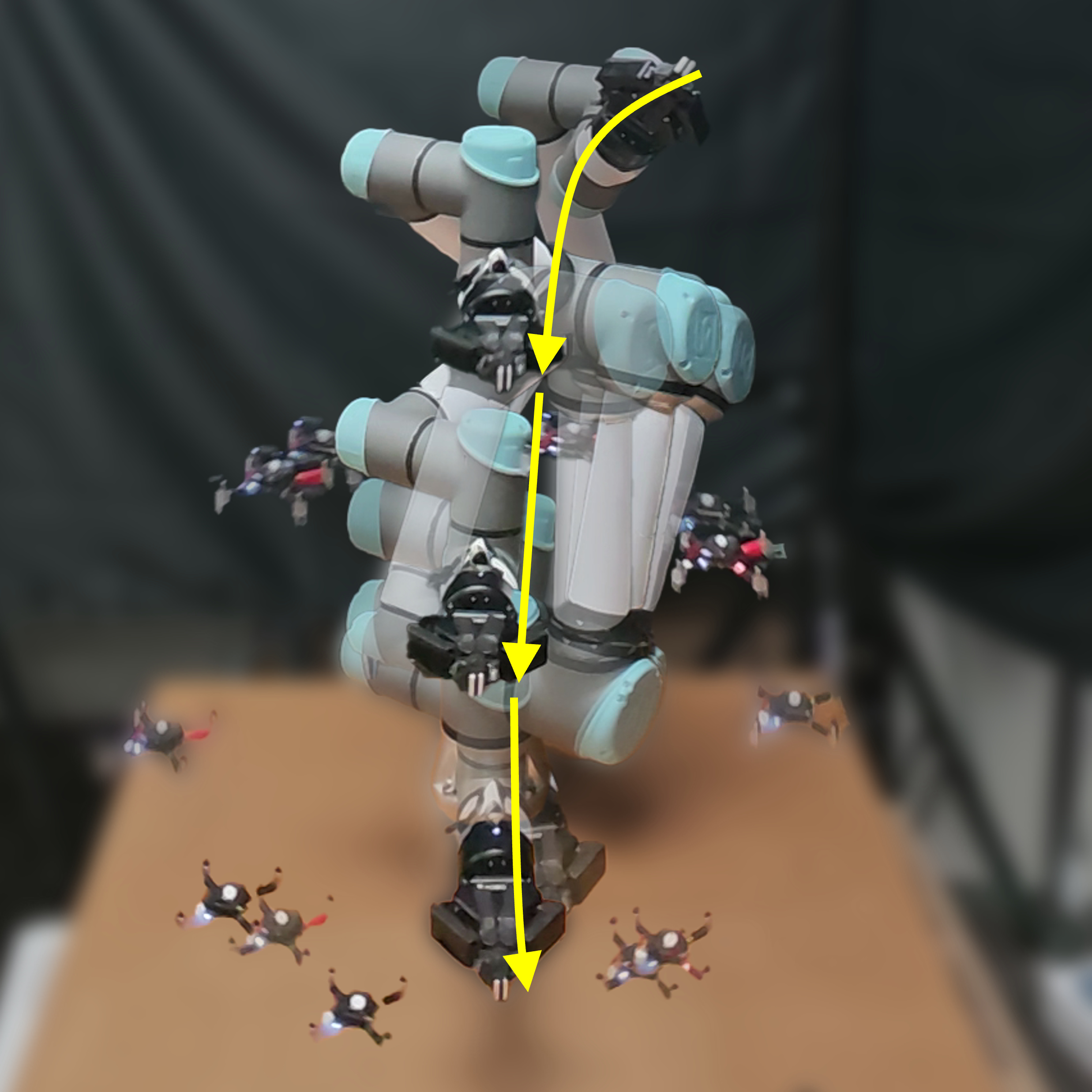}
        {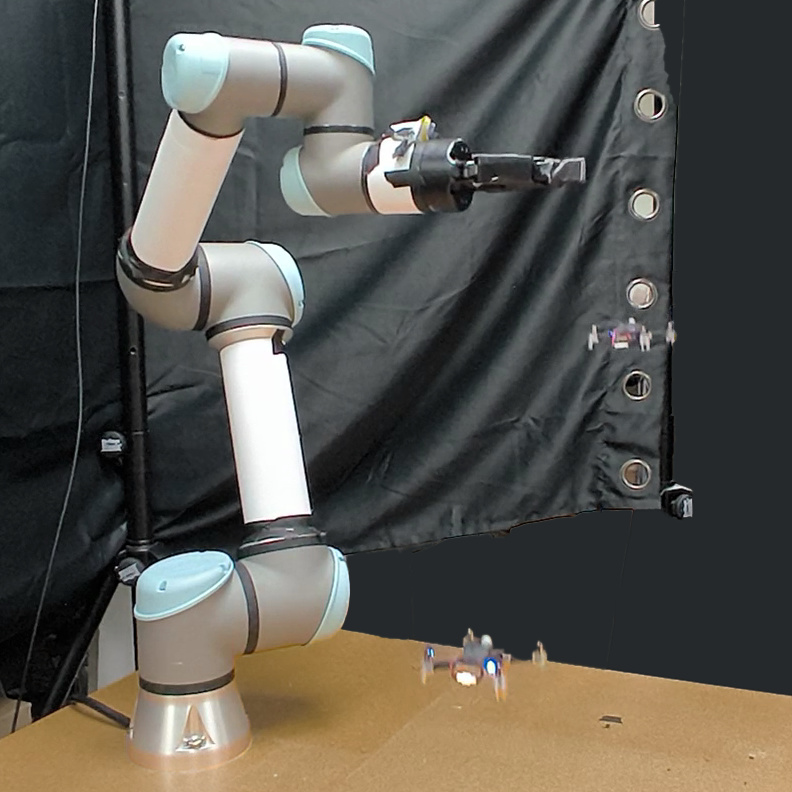}
        {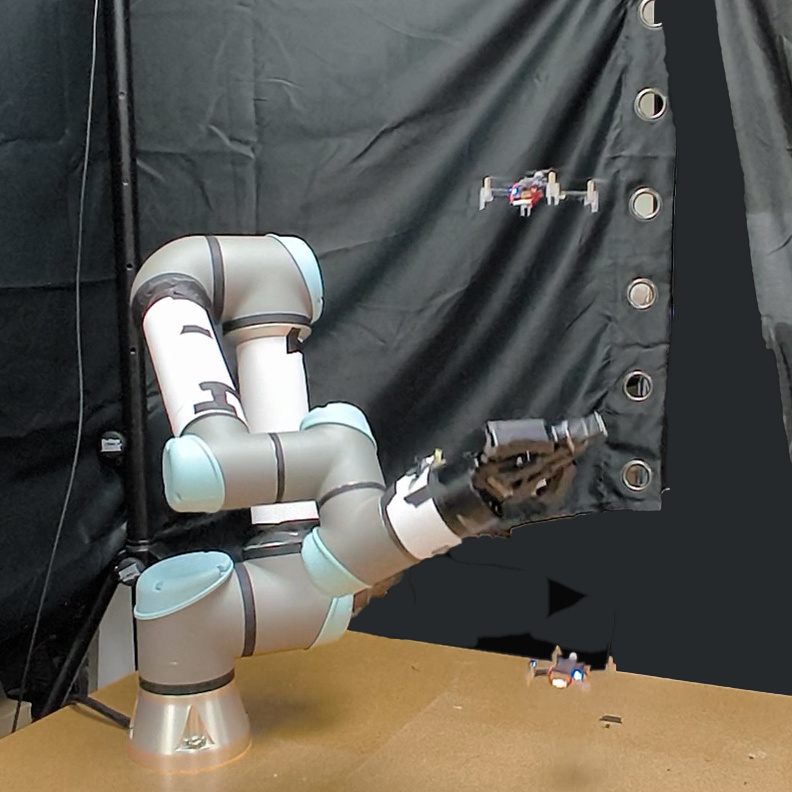}
        {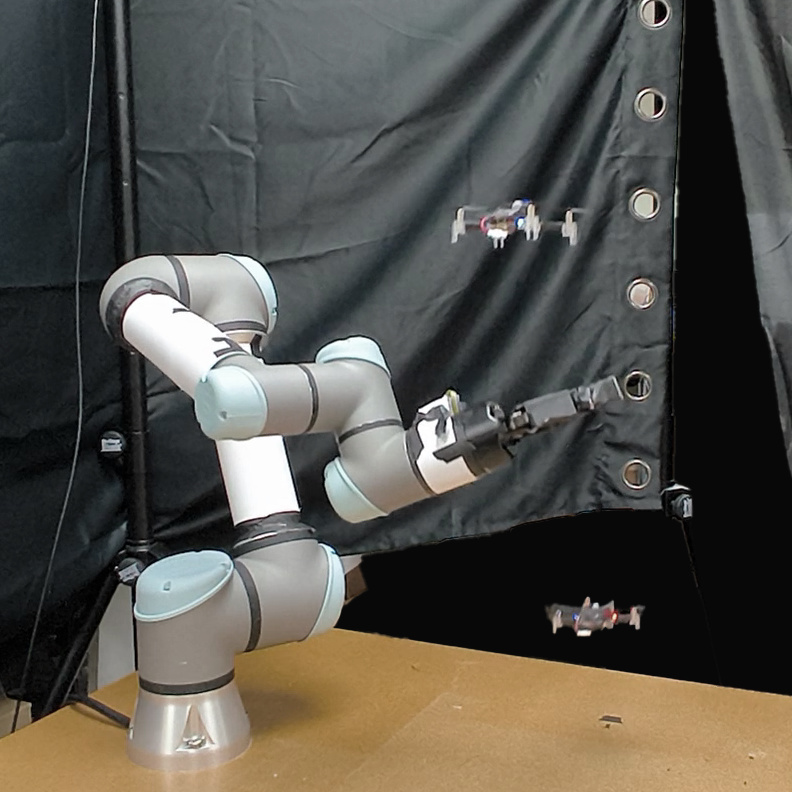}
    \end{minipage}}}%
  \hspace{.002\textwidth}
  \subfloat[Fast three-Crazyflie left--right reaching.\label{fig:hardware-fast}]{%
    \resizebox{.325\textwidth}{!}{%
    \begin{minipage}[t]{.35\textwidth}
      \centering
      \hardwaremontage{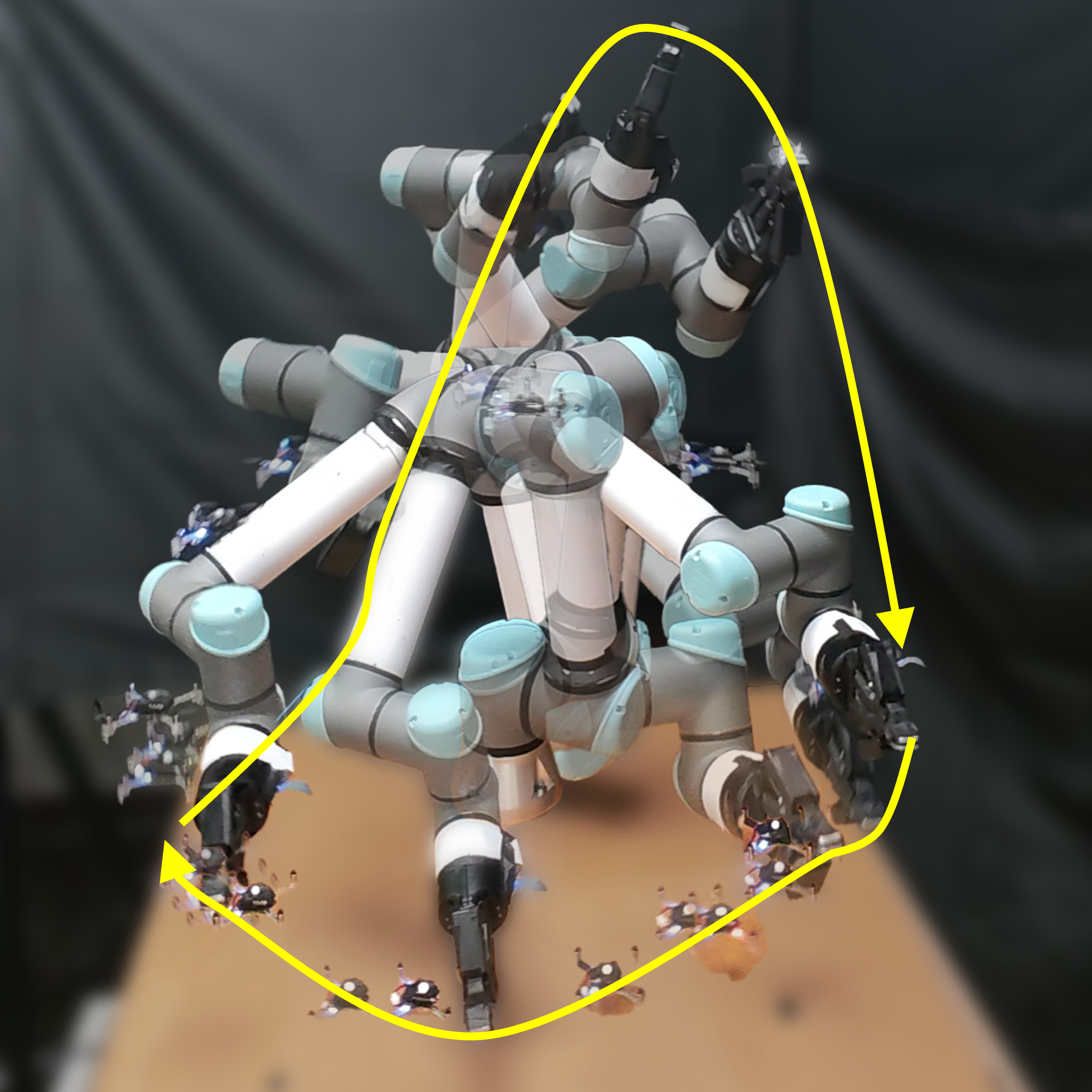}
        {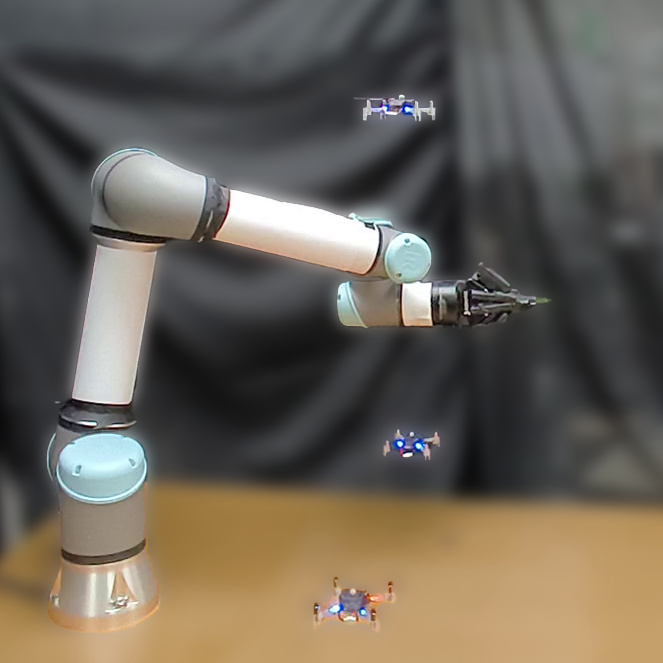}
        {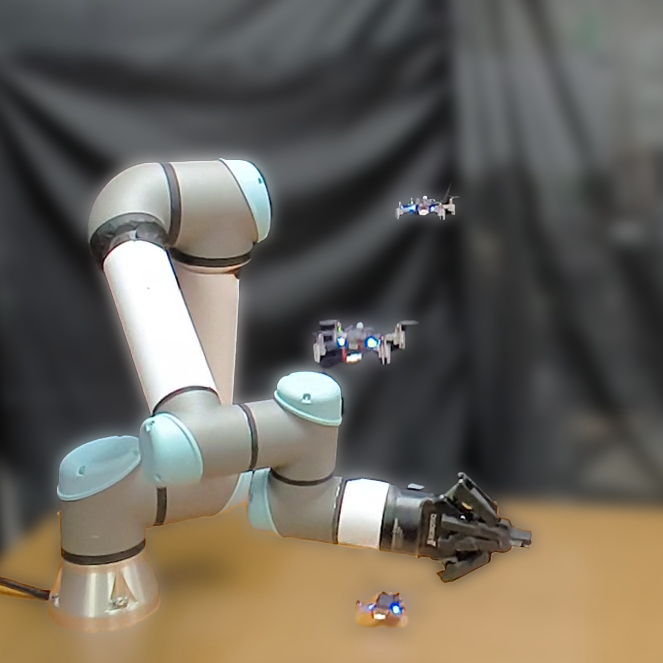}
        {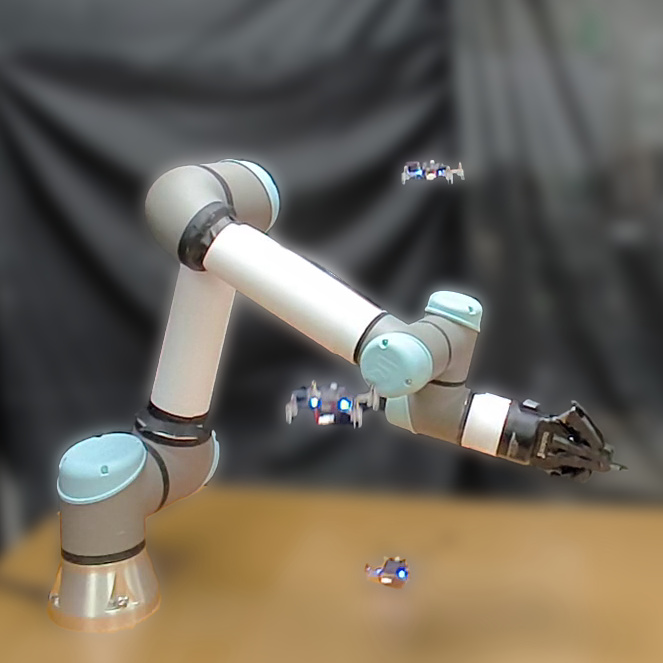}
    \end{minipage}}}%
  \hspace{.002\textwidth}
  \subfloat[Five-Crazyflie portal-poking.\label{fig:hardware-five}]{%
    \resizebox{.325\textwidth}{!}{%
    \begin{minipage}[t]{.35\textwidth}
      \centering
      \hardwaremontage{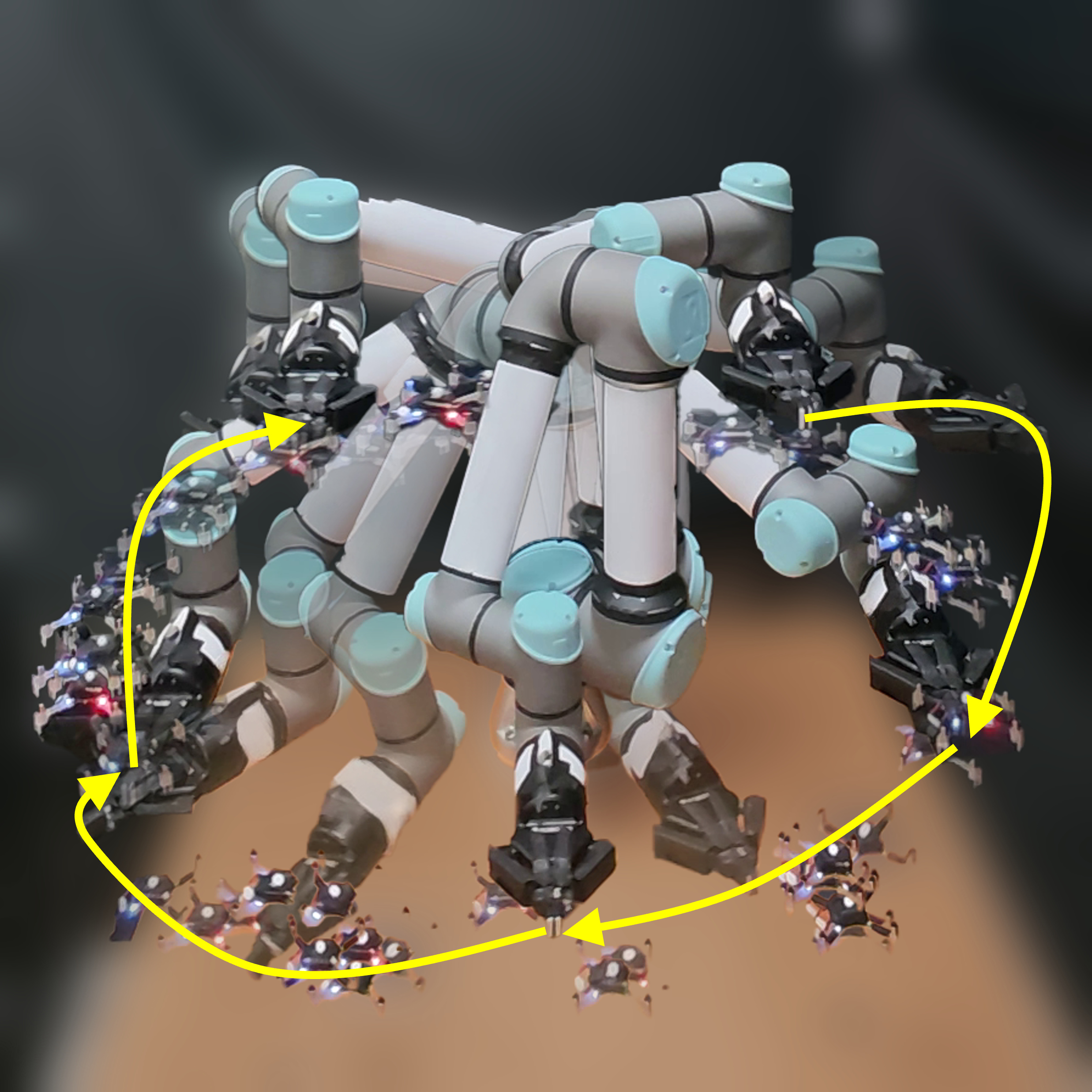}
        {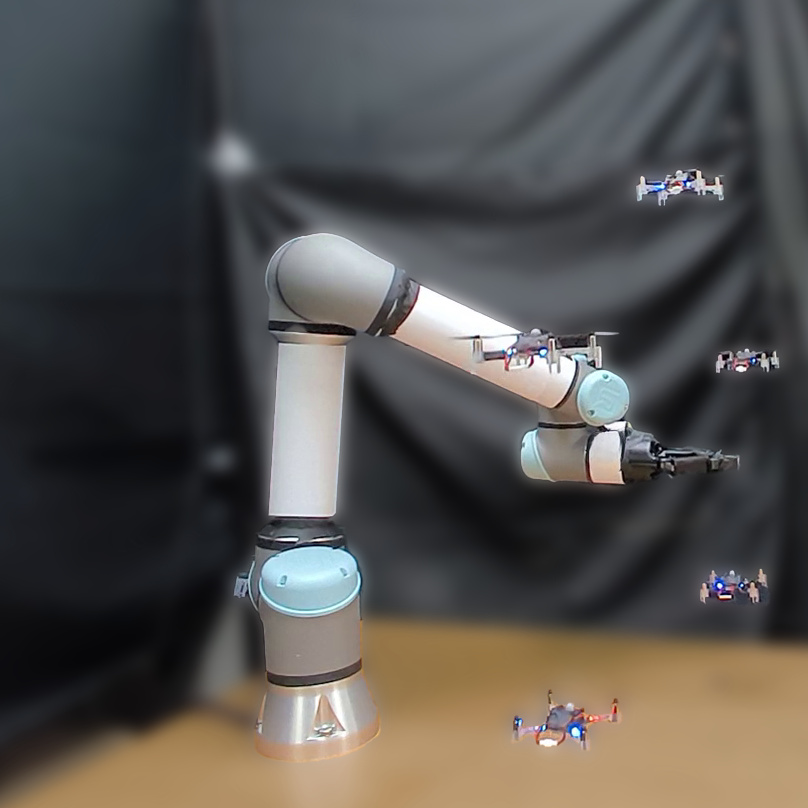}
        {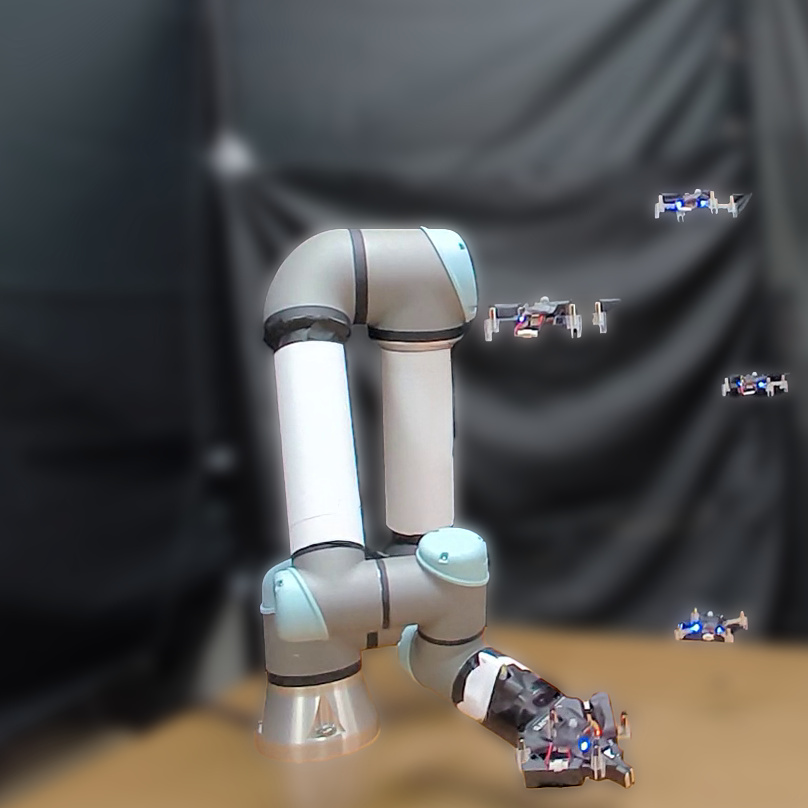}
        {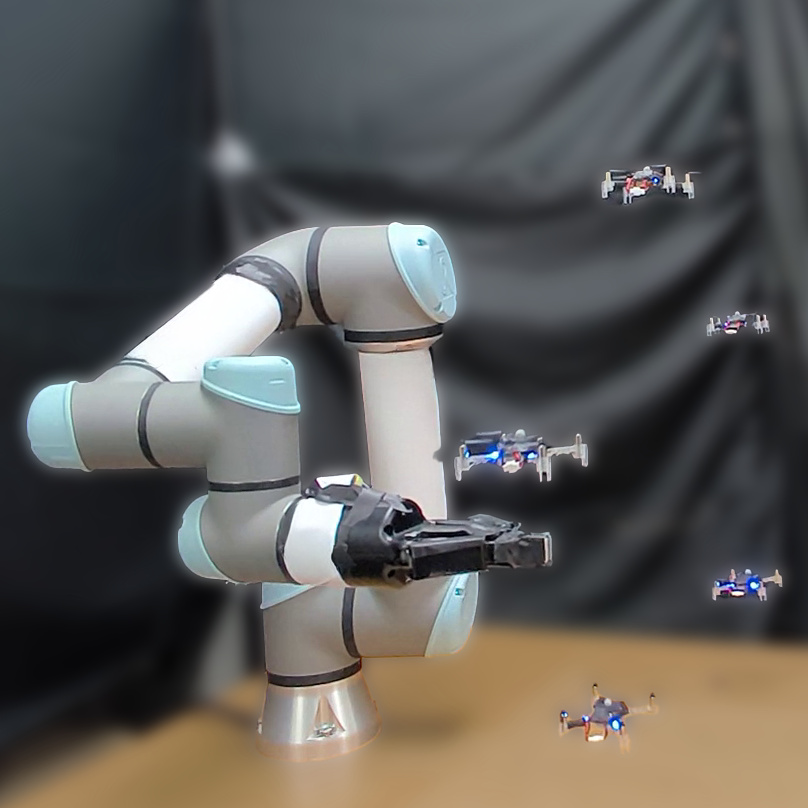}
    \end{minipage}}}
      \vspace{3mm}
  \caption{Physical executions of root-recycling \method{}. Each large image
  overlays the same UR5e and its moving drone obstacles at multiple times.
  Yellow arrows indicate end-effector motion between successive goals;
  the smaller images show selected side-camera snapshots. The slow
  three-Crazyflie demonstration is shown in Fig.~\ref{fig:hardware-intro}.}
  \label{fig:hardware}
\end{figure*}

We demonstrate root-recycling \method{} on a UR5e with a Robotiq gripper among
two, three, and five OptiTrack-localized Crazyflies. Prescribed piecewise-polynomial
drone trajectories use one conservative spherical collision proxy per vehicle.
The robot plans motions to several consecutive goals given the drones'
trajectories. After drone
prepositioning, a synchronized trigger executes the preplanned arm and drone
trajectories; no online
replanning occurs.

The slow three-Crazyflie left--right reaching demonstration appears in
Fig.~\ref{fig:hardware-intro}. Figure~\ref{fig:hardware-two} shows two Crazyflies
circling a vertical portal as the gripper reaches its top, center, and bottom
before returning home. Figures~\ref{fig:hardware-fast} and~\ref{fig:hardware-five}
show fast three-Crazyflie left--right reaching and five-Crazyflie portal-poking.
Each montage pairs a temporal overlay with selected side-camera snapshots.
Together the figures document physical execution of all four exported plans.

%% file: texs/15-conclusion.tex
ST-pRRTC couples a GPU-resident forward tree with an adaptive goal-time forest.
Interval root is probabilistically complete and asymptotically arrival-time
optimal under the stated assumptions. Root recycling has no such guarantees
but concentrates finite search effort on useful trees.
Both variants achieve lower mean first-solution times and earlier mean final
arrivals than ST-RRT* and SI-RRT on common-success sets, including across
three seeds at 5~s. Root recycling also achieves the highest tabletop success
and maintains early arrivals with sparse roots over broad horizons.
UR5e demonstrations show execution among moving Crazyflies.